\documentclass[runningheads]{llncs}

\usepackage[utf8]{inputenc}
\usepackage{float}
\usepackage{graphicx}
\usepackage{multirow}
\usepackage{amsmath}
\usepackage{xcolor}
\usepackage{amssymb}
\usepackage{latexsym}
\usepackage{booktabs}
\usepackage{arydshln}
\usepackage{bibnames}
\usepackage{wrapfig}
\usepackage[pagebackref,breaklinks,colorlinks]{hyperref}
\usepackage[capitalize]{cleveref}

\UseRawInputEncoding

\crefname{section}{Sec.}{Secs.}
\crefname{table}{Tab.}{Tabs.}
\crefname{figure}{Fig.}{Figs.}

\newcommand{\Footnotemark}{\footnotemark%
 \expandafter\global\expandafter\let\csname saved@Href@\alphalph{\value{footnote}}\endcsname%
    \Hy@footnote@currentHref}
\newcommand{\Footnotetext}[1]{%
   \expandafter\let\expandafter\Hy@footnote@currentHref\csname saved@Href@\alphalph{\value{footnote}}\endcsname%
   \footnotetext{#1}}
\makeatother

\def\our{StyleAE }
\def\nor{\mathcal{N}}
\def\R{\mathbb{R}}

\begin{document}

\title{StyleAutoEncoder for manipulating image attributes using pre-trained StyleGAN}
\titlerunning{StyleAutoEncoder}

\author{Andrzej Bedychaj\orcidID{0000-0001-9416-3991} \and
Jacek Tabor\orcidID{0000-0001-6652-7727} \and
Marek \'Smieja\orcidID{0000-0003-2027-4132}}
\authorrunning{A. Bedychaj et al.}
%
\institute{Faculty of Mathematics and Computer Science, Jagiellonian University, Kraków, Poland\\
\email{andrzej.bedychaj@student.uj.edu.pl \\marek.smieja@uj.edu.pl \\}}

\maketitle
\setcounter{footnote}{0}

\begin{abstract}
Deep conditional generative models are excellent tools for creating high-quality images and editing their attributes. However, training modern generative models from scratch is very expensive and requires large computational resources. In this paper, we introduce StyleAutoEncoder (StyleAE), a lightweight AutoEncoder module, which works as a plugin for pre-trained generative models and allows for manipulating the requested attributes of images. The proposed method offers a cost-effective solution for training deep generative models with limited computational resources, making it a promising technique for a wide range of applications. We evaluate \our by combining it with StyleGAN, which is currently one of the top generative models. Our experiments demonstrate that \our is at least as effective in manipulating image attributes as the state-of-the-art algorithms based on invertible normalizing flows. However, it is simpler, faster, and gives more freedom in designing neural architecture.
\end{abstract}

\begin{figure}[]
\vspace{-2em}
\centering
\scalebox{0.95}{
\tiny
\begin{tabular}{cccccccccc}
\textbf{Input} & \textbf{Old}  & \textbf{Beard} & \textbf{Gender} & & & \textbf{Input} & \textbf{No Glasses} & \textbf{Young} & \textbf{Smile}  \\
\includegraphics[width=0.12\linewidth]{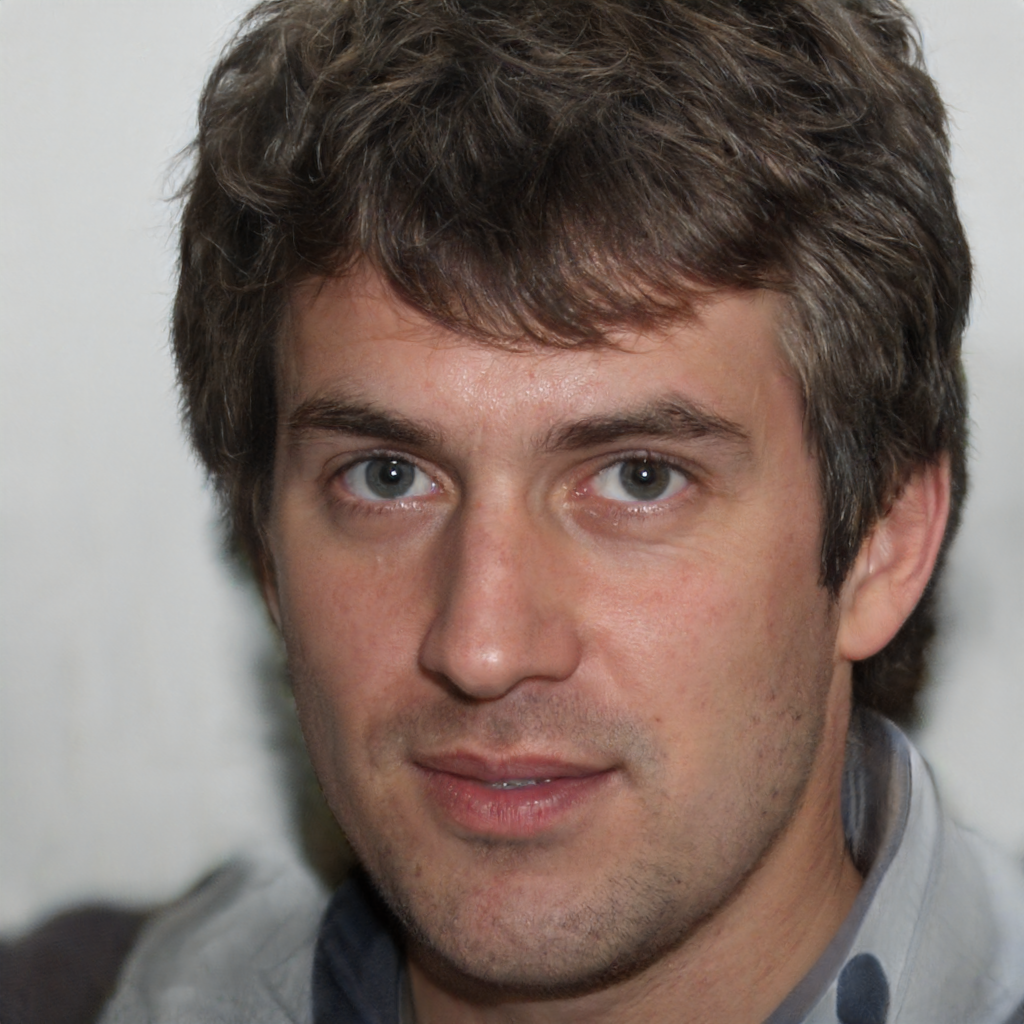} & \includegraphics[width=0.12\linewidth]{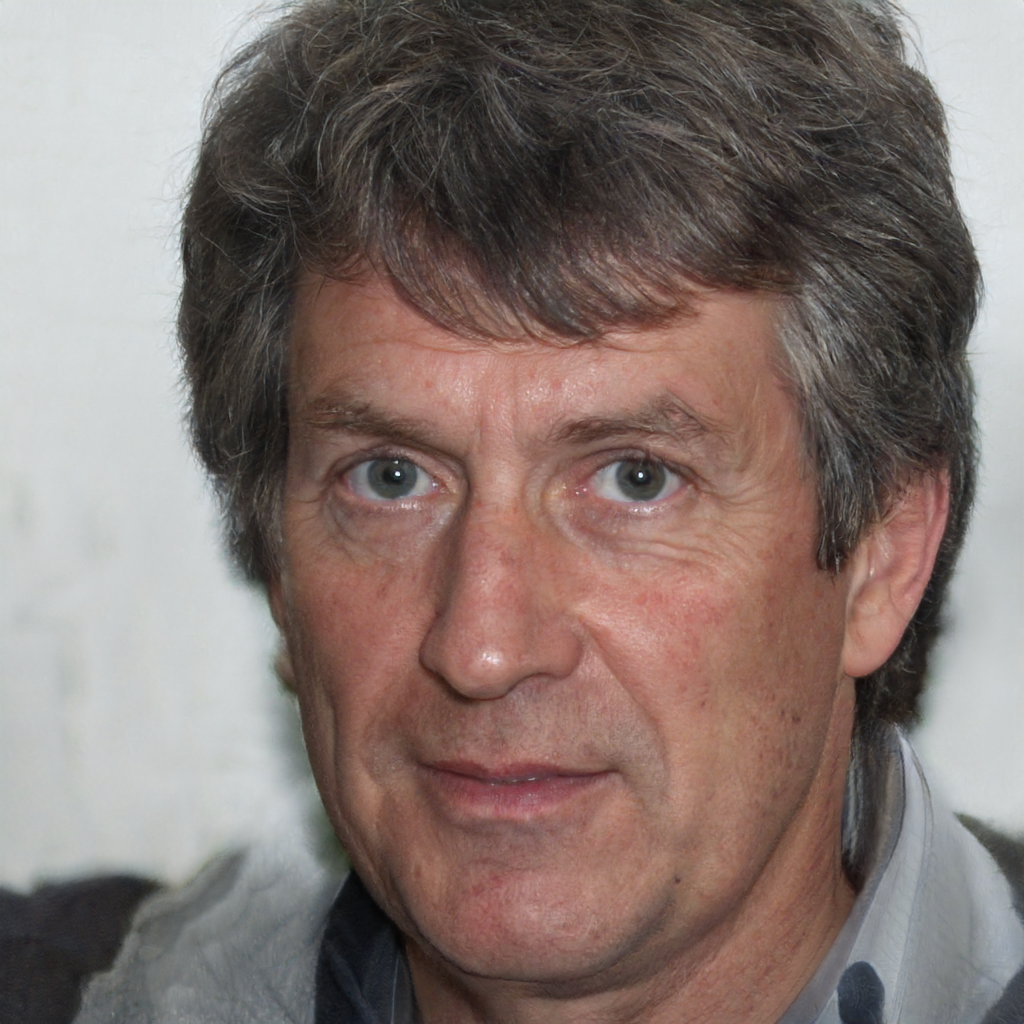}  & \includegraphics[width=0.12\linewidth]{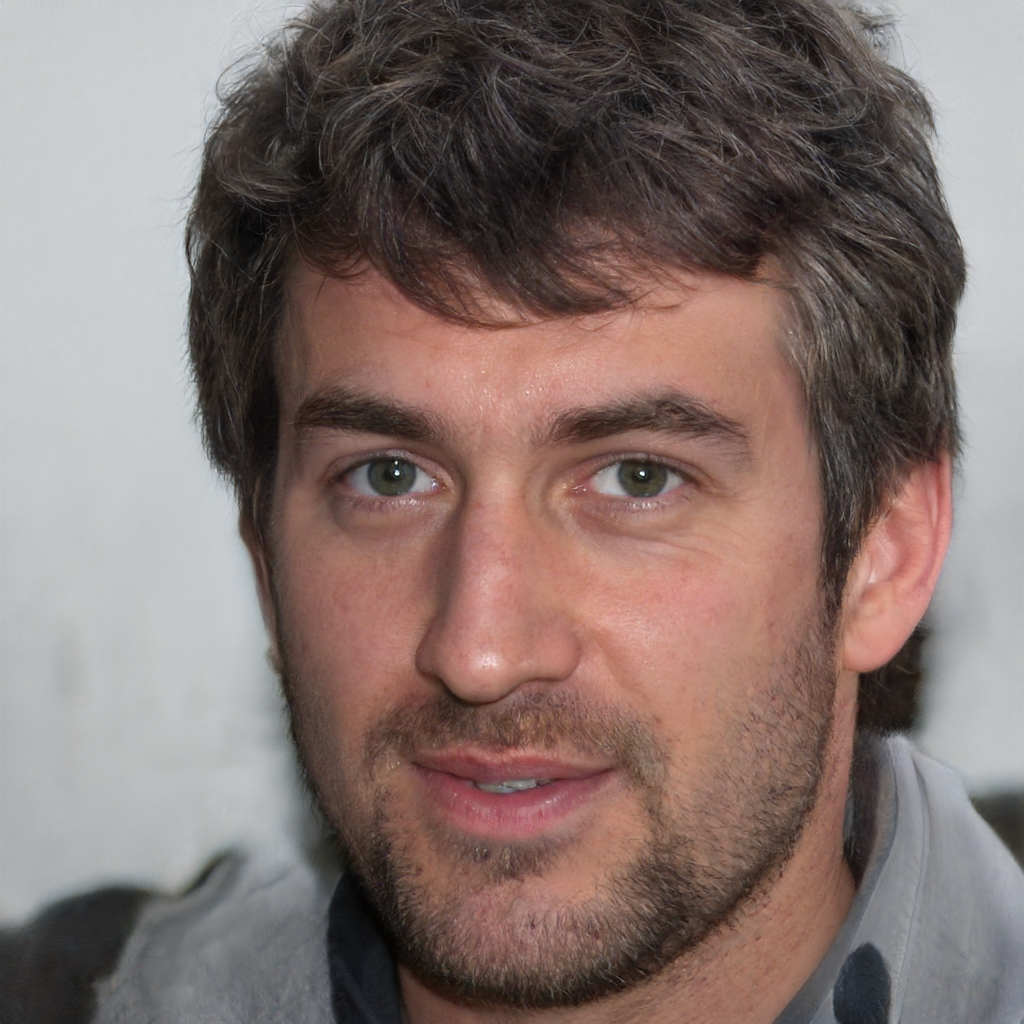}  & 
\includegraphics[width=0.12\linewidth]{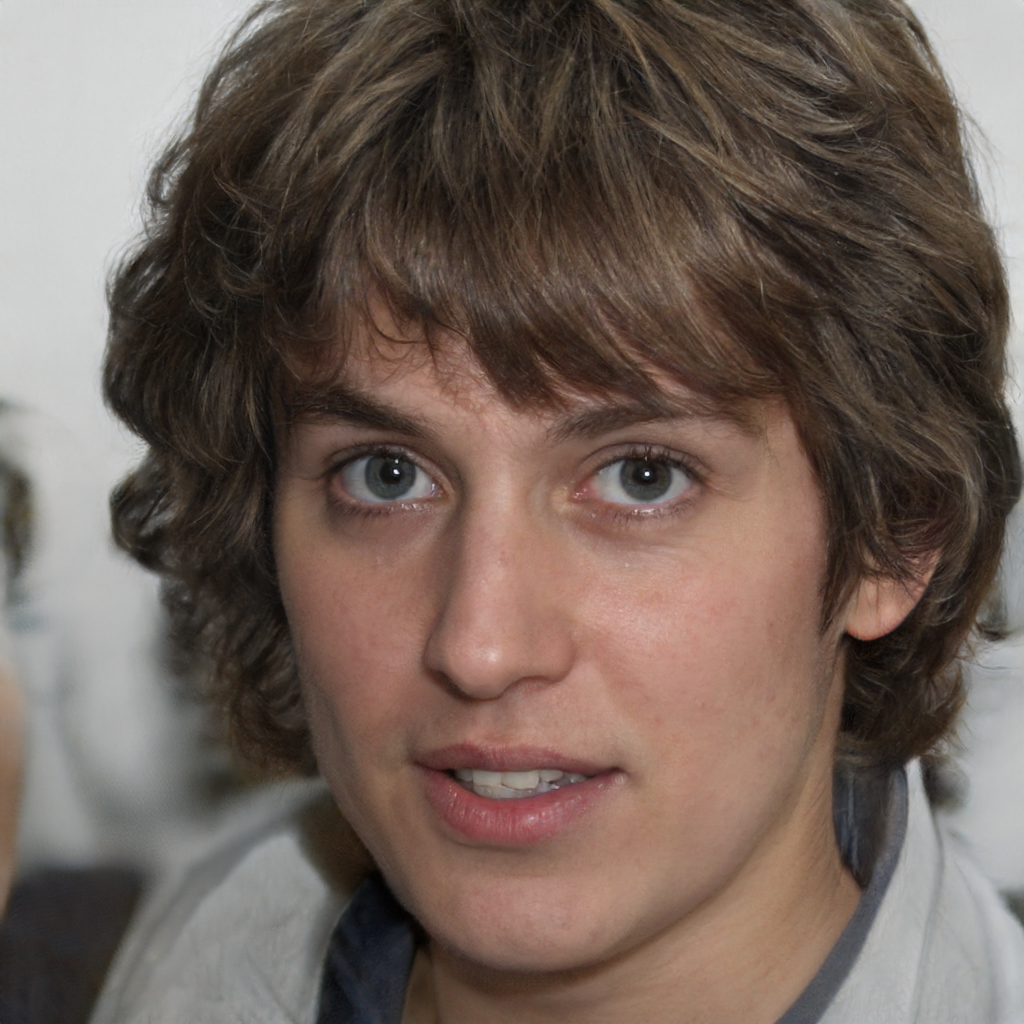} & 
&
&
\includegraphics[width=0.12\linewidth]{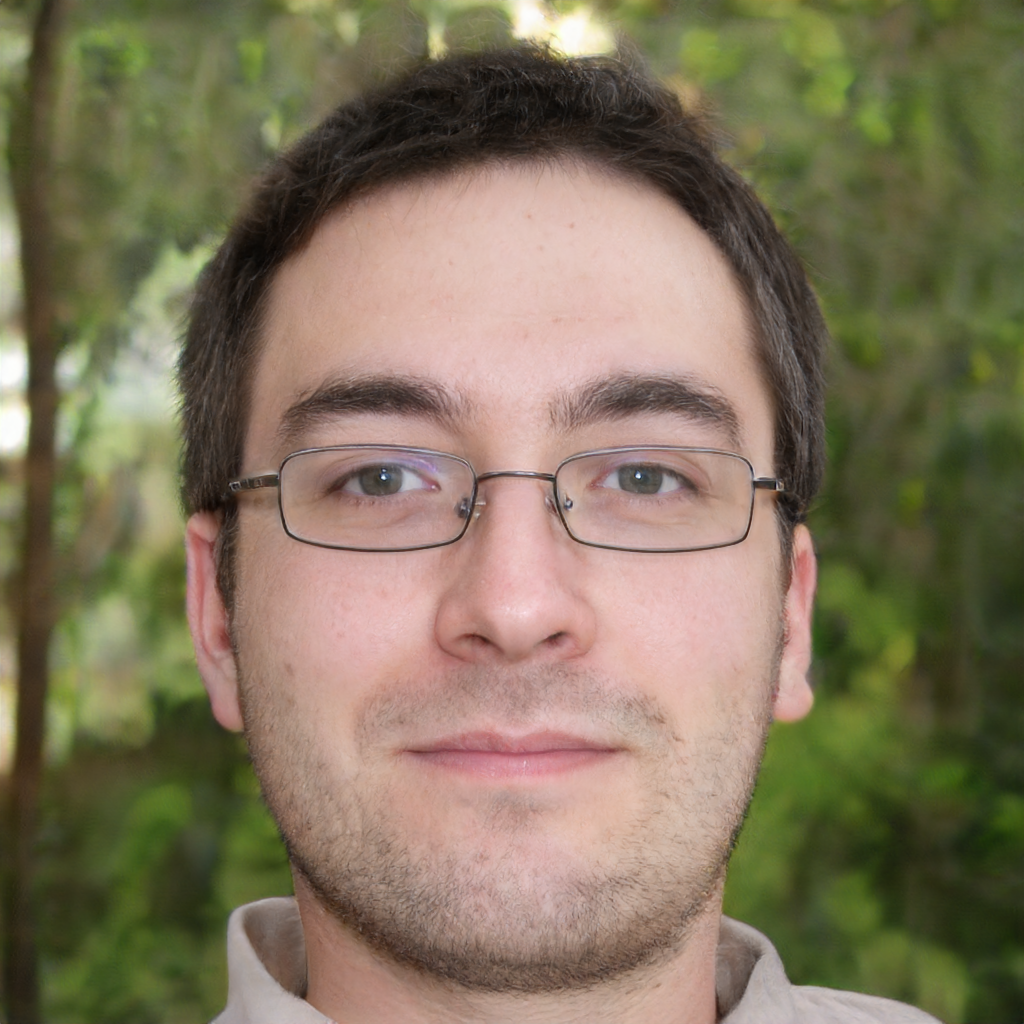} &
\includegraphics[width=0.12\linewidth]{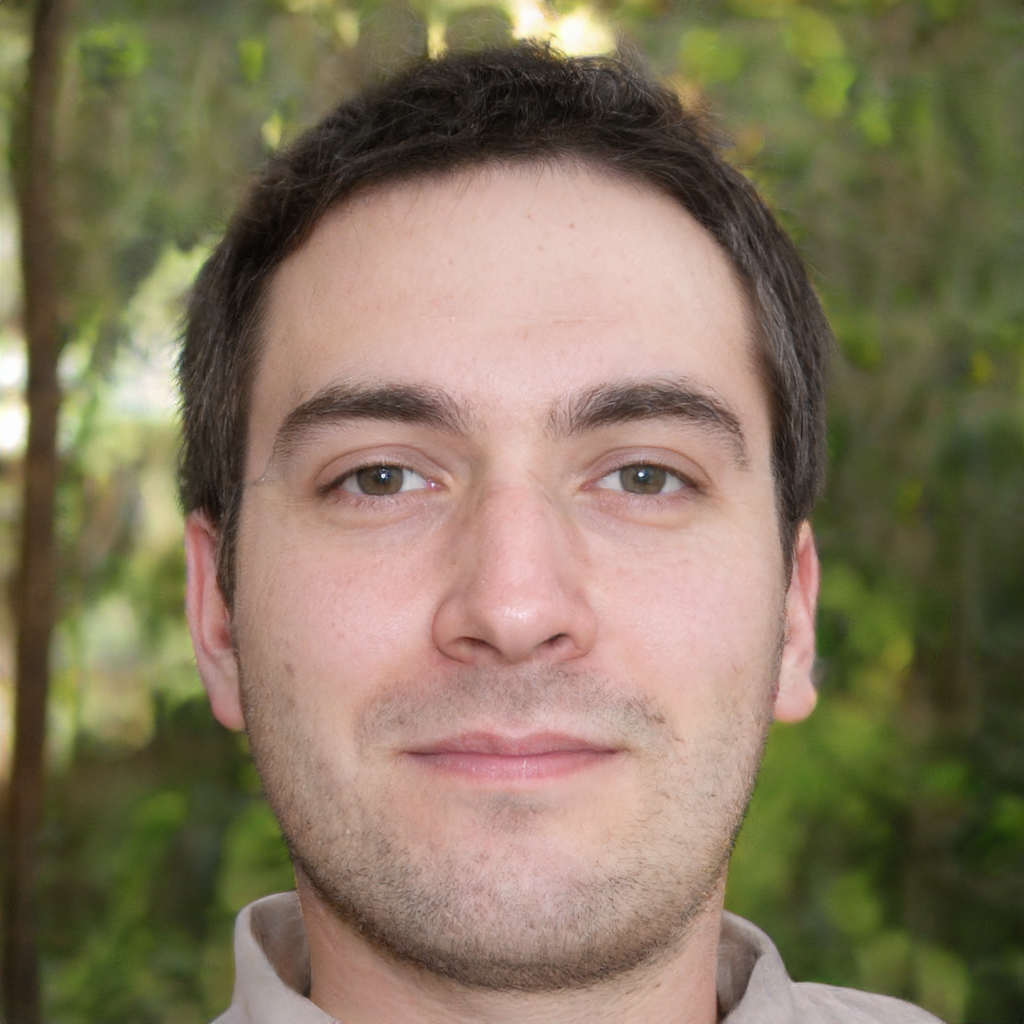} & 
\includegraphics[width=0.12\linewidth]{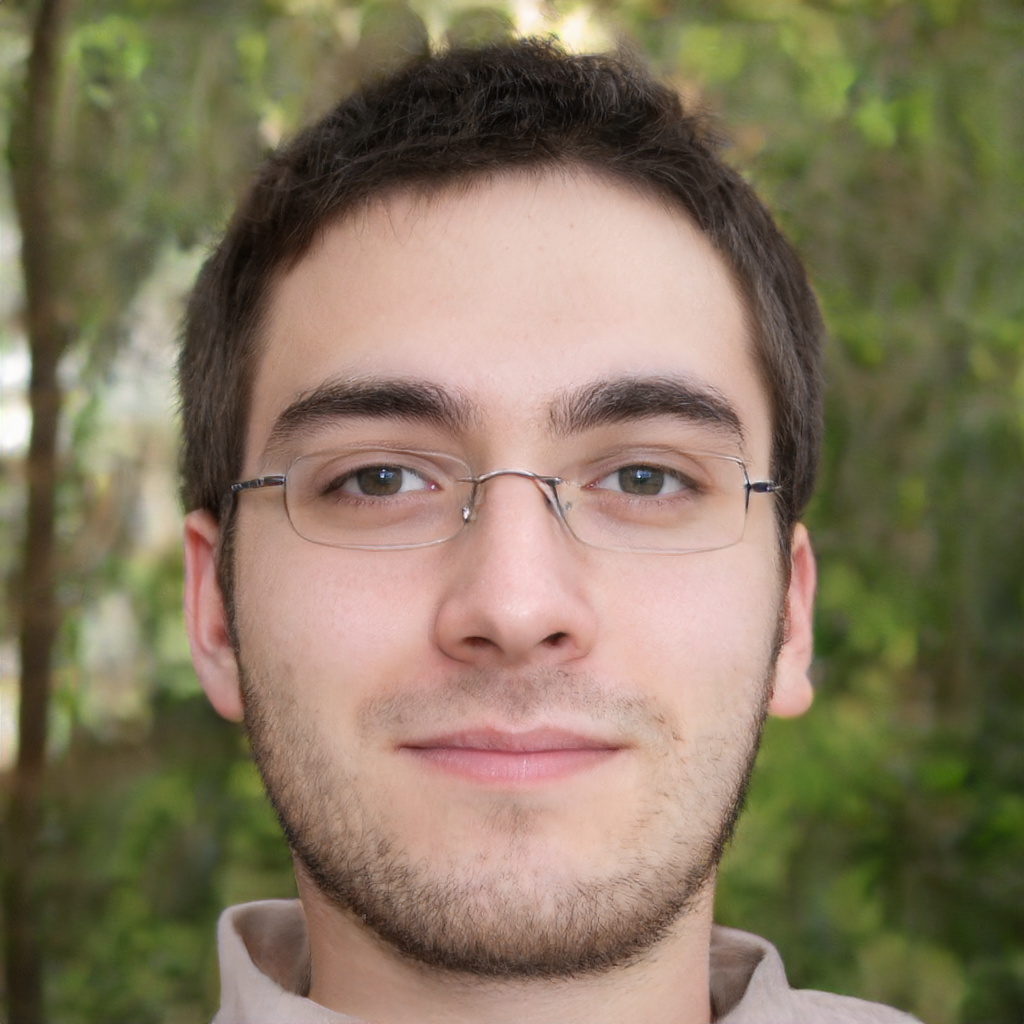} &
\includegraphics[width=0.12\linewidth]{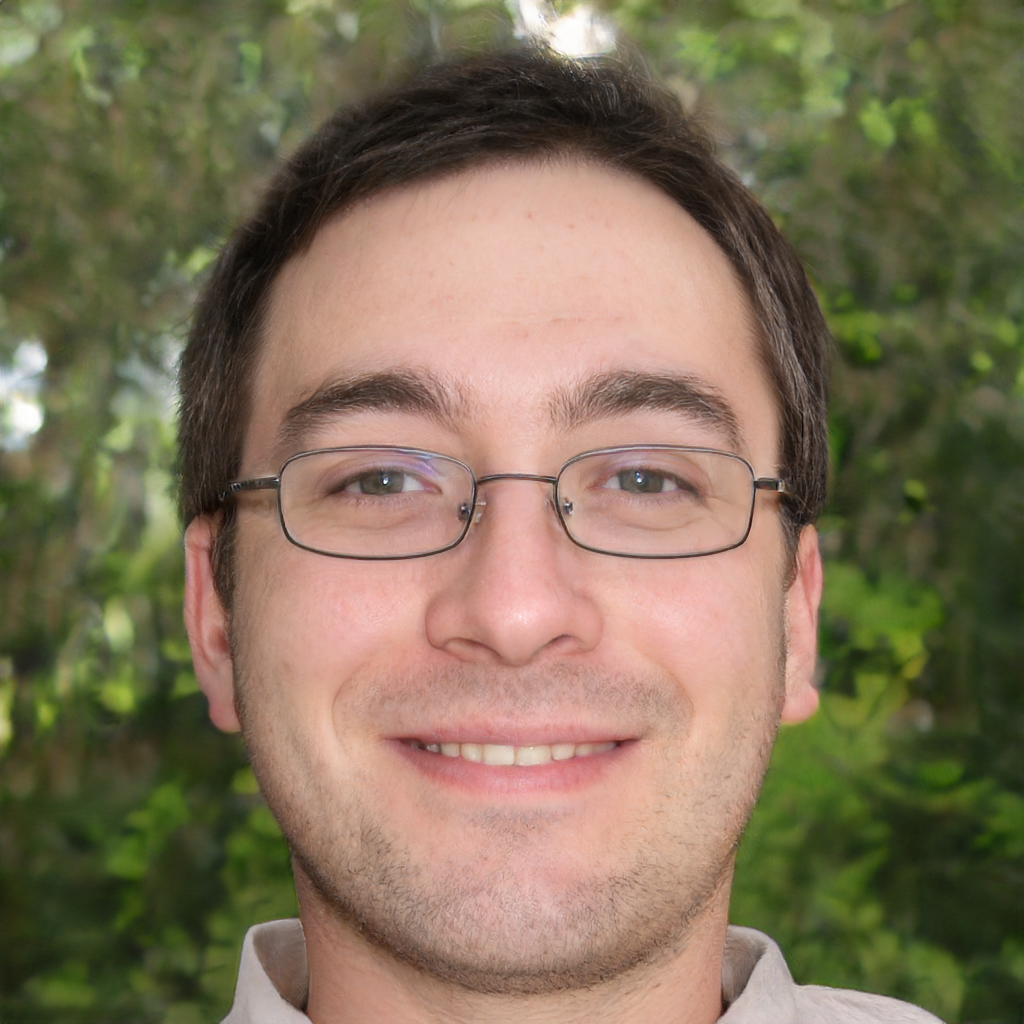} 
\end{tabular}%
}
\caption{Single-attribute manipulation with StyleAE in StyleGAN's latent space.}
\label{showcase}
\vspace{-3em}
\end{figure}

\section{Introduction}

Generative models, such as generative adversarial networks (GAN)~\cite{goodfellow2014generative}, variational AutoEncoders (VAE)~\cite{kingma2022autoencoding}, diffusion models~\cite{preechakul2022diffusion}, and flow-based generative models \cite{dinh2017density}, have gained popularity due to their ability to create high-quality images, videos, and texts. These models are trained using supervised or unsupervised learning techniques on large datasets. They have transformed the field of artificial intelligence and are used to create innovative applications across various research fields \cite{he2019attgan,wang2020hijack,tewari2020pie,shen2020interfacegan,liu2019conditional,gao2021high}.

StyleGAN \cite{karras2019stylebased,karras2020analyzing,karras2021aliasfree} is one of the most popular generative models used for creating high-quality images, known for its ability to control various aspects of the generated image such as pose, expression, and style. However, the latent space of StyleGAN is highly entangled~\cite{karras2019stylebased}, meaning that the different attributes of the generated image are not easily separable. This makes it challenging to manipulate individual attributes without affecting others, limiting the controllability of the generated images. Furthermore, manipulating the latent space of StyleGAN is a challenging task due to the complex and high-dimensional nature of the model, which limits its practical applications. 

There are various methods for simplifying the latent space of generative models. Unsupervised methods include algorithms such as Interface GAN~\cite{shen2020interfacegan}, GANSpace~\cite{harkonen2020ganspace} and InfoGAN~\cite{chen2016infogan}, which aim to learn a disentangled representation of the data without requiring any labeled data. Supervised methods like PluGeN~\cite{wolczyk2022plugen}~\cite{suwala2024face} or StyleFlow \cite{styleflow}, on the other hand, require labeled data and involve training an auxiliary model to predict a particular attribute, such as the pose or shape of the generated image. These approaches are essential for improving the practical applications of generative models and making them more useful for a wider range of applications.

While flow-based models such as StyleFlow and PluGeN have shown promising results in disentangling the latent space of StyleGAN, they also have some limitations. One significant limitation is the difficulty in scaling these models to high-resolution images due to their computationally intensive nature \cite{kingma2018glow}. Moreover, invertible models require a large amount of training data to learn the complex data distributions \cite{ho2019flow}, which may be challenging to obtain in some cases. Finally, flow-based models can be sensitive to hyperparameters \cite{vidal2022taming}, making them difficult to develop optimal performance.

This paper presents a novel approach, called StyleAE, for modifying image attributes using a combination of AutoEncoders with StyleGAN. StyleAE simplifies the~latent space of StyleGAN so that the values of target attributes can be effectively changed. While StyleAE achieves at least as good results as the state-of-the-art flow-based methods, it is computationally more efficient and does not require so large amount of training data, which makes it a practical approach for various applications (see \cref{showcase} for sample results). 

We conducted an assessment of \our through extensive experiments on datasets containing images of human and animal faces, benchmarking it against state-of-the-art flow-based models. Our findings demonstrate that StyleAE's effectiveness in manipulating the latent space of StyleGAN is on par with that of flow-based models. Our research provides crucial insights into the unique strengths and limitations of \our model, highlighting the potential of our algorithm to improve the effectiveness of latent space manipulation for StyleGAN and other generative models. Furthermore, our approach exhibits superior time complexity, making it a more feasible solution for a wide range of applications.

\section{Related Work}
Conditional VAE (cVAE) introduced label information integration into generative models but lacks assured latent code and label independence, impacting generation quality; on the other hand, Conditional GAN (cGAN) produces higher-quality outputs but involves more complex training and falls short in manipulating existing examples \cite{kingma2014semi,he2019attgan}.

Fader Networks \cite{lample2017fader} address this limitation by combining cVAE and cGAN components, utilizing an encoder-decoder architecture and a discriminator predicting attributes. However, Fader Networks struggle with attribute disentanglement, and their training is more challenging than standard GANs. CAGlow \cite{liu2019conditional} takes a different approach, using Glow for conditional image generation based on joint probabilistic density modelling. However, it does not reduce data dimensionality, limiting its applicability to more complex data. Competitive approaches like HifaFace \cite{gao2021high}, Pie \cite{tewari2020pie}, and GANSpace \cite{harkonen2020ganspace} have also been explored.

InterFaceGAN \cite{shen2020interfacegan} and Hijack-GAN \cite{wang2020hijack} manipulate facial semantics through linear models and a proxy model for latent space traversal. Recent approaches focus on manipulating latent codes of pre-trained networks, where data complexity is less restrictive, making flow models applicable. StyleFlow \cite{styleflow} and PluGeN \cite{wolczyk2022plugen} use normalizing flow modules on GAN latent spaces, employing conditional CNF and NICE, respectively. While StyleFlow is tailored for StyleGAN, PluGeN achieves important results across various models and domains. Further extensions of PluGeN on face images (dubbed PluGeN4Faces) \cite{suwala2024face} allowed for a significant improvement in the disentanglement between the attributes of the face and the identity of the person.

Our work takes a distinct approach, achieving results comparable to these methods while providing superior attribute decomposition and structural consistency for image datasets, coupled with significantly improved computational efficiency.

\section{Methodology}

In this section, we give a description of the proposed \our{} approach to disentangling the latent space of generative models. Before that, we recall the StyleGAN and AutoEncoder architectures, which are the main building blocks of \our{}.

\subsection{Preliminaries}

\paragraph{StyleGAN\cite{karras2019stylebased,karras2020analyzing,karras2021aliasfree}:} is a cutting-edge generative model lauded for its capacity to produce high-quality, realistic images. Its architecture comprises two key elements. Initially, the latent vector $z$, generated from a standard Gaussian distribution $\nor(0,I)$, undergoes mapping to the style space vector $w$ through a series of fully connected layers. Subsequently, this style vector is injected into the following convolutional blocks of the synthesis network, progressively crafting the image in the desired resolution.

While the latent vector $z$ serves as the foundation for image creation, manipulating the image via the style vector $w$ is notably more convenient. As the replicated style vector influences various blocks of the synthesis network, it enables the control of the image's style at different abstraction levels, offering users versatile means to manipulate generated images. However, achieving control over specific attributes without impacting others requires disentangling the style space. In the subsequent sections, we detail how we employ an AutoEncoder to modify the latent space of StyleGAN, enhancing its separability and controllability.

\begin{wrapfigure}{r}{0.5\textwidth}
\vspace{-3em}
\includegraphics[width=0.48\textwidth]{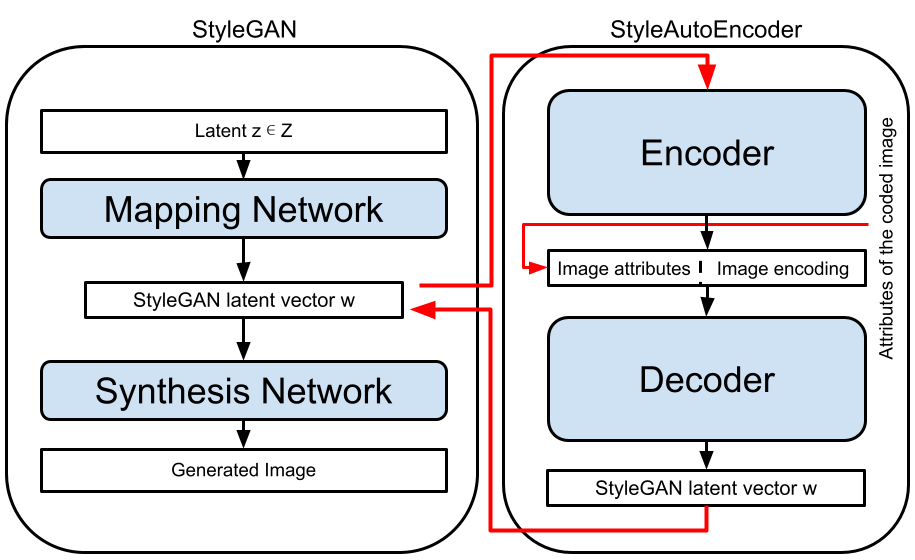}
\caption{ \textbf{Architecture design of StyleAE.} \our{} maps the style code $w$ of the pre-trained StyleGAN into a target space, where labelled attributes are modelled by individual coordinates. 
\label{diagram}}
\vspace{-1em}
\end{wrapfigure}

\paragraph{AutoEncoder:} architecture comprises an encoder function, mapping input data to a compressed representation in the latent space, and a decoder function, mapping this representation back to the original input space. This is represented as:
\begin{equation}
z = \mathcal{E}(x) \text{ and }
x'=\mathcal{D}(z),
\end{equation}
where $\mathcal{E}$ is the encoder, $\mathcal{D}$ is the decoder, $x$ is the input data, $z$ is the latent space representation, and $x'$ is the reconstructed output.

Training involves minimizing the reconstruction error, typically mean squared error (MSE) or binary cross-entropy~(BCE), between $x$ and $x'$. If the latent space's dimensionality is much lower than the input resolution, the AutoEncoder learns essential features in the compressed representation $z$. In our research, as we do not emphasize image compression, we do not reduce the dimensionality.

\subsection{StyleAutoEncoder} \label{StyleAE}

Our goal is to simplify and enhance the manipulation of image attributes by modifying the latent space of StyleGAN using an AutoEncoder. To this end, we develop StyleAE, an AutoEncoder plugin to StyleGAN, which allows for convenient manipulation of the requested attributes and preserving the quality of StyleGAN, see~\Cref{diagram}.

\paragraph{Structure of the target space:} We assume that every instance $x \in X$ is associated with a $K$ dimensional vector of labels $y~=~(y_1,\ldots, y_K)$. The labels can represent a combination of discrete and continuous features. Our objective is to find a representation of images, where each label is encoded as a separate coordinate. More precisely, let the $k$-th latent variable $c_k$ correspond to the attribute $y_k$. By modifying the value of $c_k$, we would like to change the value of $k$-th attribute in the image. Since labels do not fully describe the image, additional $M$ variables $(s_1,\ldots,s_{M})$ are included to encode the remaining information. Therefore, the latent vector of our new target space is defined as $(c_1, \ldots, c_K, s_1, \ldots, s_M)$. 

To construct such a target space, we operate on the representation given by a pre-trained generative model. Although our approach theoretically applies to arbitrary generative models, we consistently use StyleGAN as a base model and fix our attention to synthesis network $G: W \to \R^N$, which maps style vectors to the images. We focus on finding a mapping between the style space $W$ and the target space $(C,S)$, where $C=(C_1, \ldots, C_k)$ describes labelled attributes and $S=(S_1, \ldots, S_M)$ denotes the remaining variables. 

\paragraph{Loss function:} To establish an approximately invertible mapping, we use an AutoEncoder-inspired neural network dubbed StyleAE. More precisely, the encoder $\mathcal{E}:~W~\to~(C,S)$~~focuses on retrieving the attributes from the style vector while the decoder  $\mathcal{D}:~ (C,S)~\to~W$ is used to recover the input data. \our{} applies the cost function, which consists of two components: attribute loss and image loss. 

To explain the details behind our loss, let $w$ be the style vector representing the image $x$ with attributes $y$, $(c,s)=\mathcal{E}(w)$ be the target representation, and $\hat{w} = \mathcal{D}(c,s)$ be the recovered style code. The image loss
\begin{equation}
    d_I(x, G(\hat{w})) = \|x - G(\hat{w})\|^2
\end{equation}
aims at reconstructing the image from AutoEncoder representation while the attribute loss
\begin{equation}
    \sum_{k=1}^K d_A(c_k,y_k),
\end{equation}
aligns target coordinates $c_k$ with image attributes. While the mean-square error (MSE) is the obvious choice for implementing attribute loss $d_A$, we found that for binary attributes $y_k \in \{0,1\}$ an alternative loss can be beneficial. Namely, for positive label $y_k=1$, we calculate the distance between the value $c_k$ returned by AE and the interval $[y_k,\infty)=[1,\infty)$ as follows:

\begin{equation}
    d_A^S(c_k, 1)= \max(0,1-c_k) \label{animals-loss}
\end{equation}
This allows us to encode a different style for a binary value, e.g.~different type of facial hair. For negative label $y_k=0$, we use typical MSE since this corresponds to the absence of an attribute.

\begin{figure}[]
\centering
\scalebox{0.9}{
\begin{tabular}{ccccccccccc}
 & \textbf{Input} & \textbf{\our}  & \textbf{StyleFlow} & \textbf{PluGeN} &  & & \textbf{Input} & \textbf{\our}  & \textbf{StyleFlow} & \textbf{PluGeN} \\
 \multirow{1}{*}[7.5ex]{\rotatebox[origin=c]{90}{\textbf{glasses}}} & 
 \includegraphics[width=0.12\linewidth]{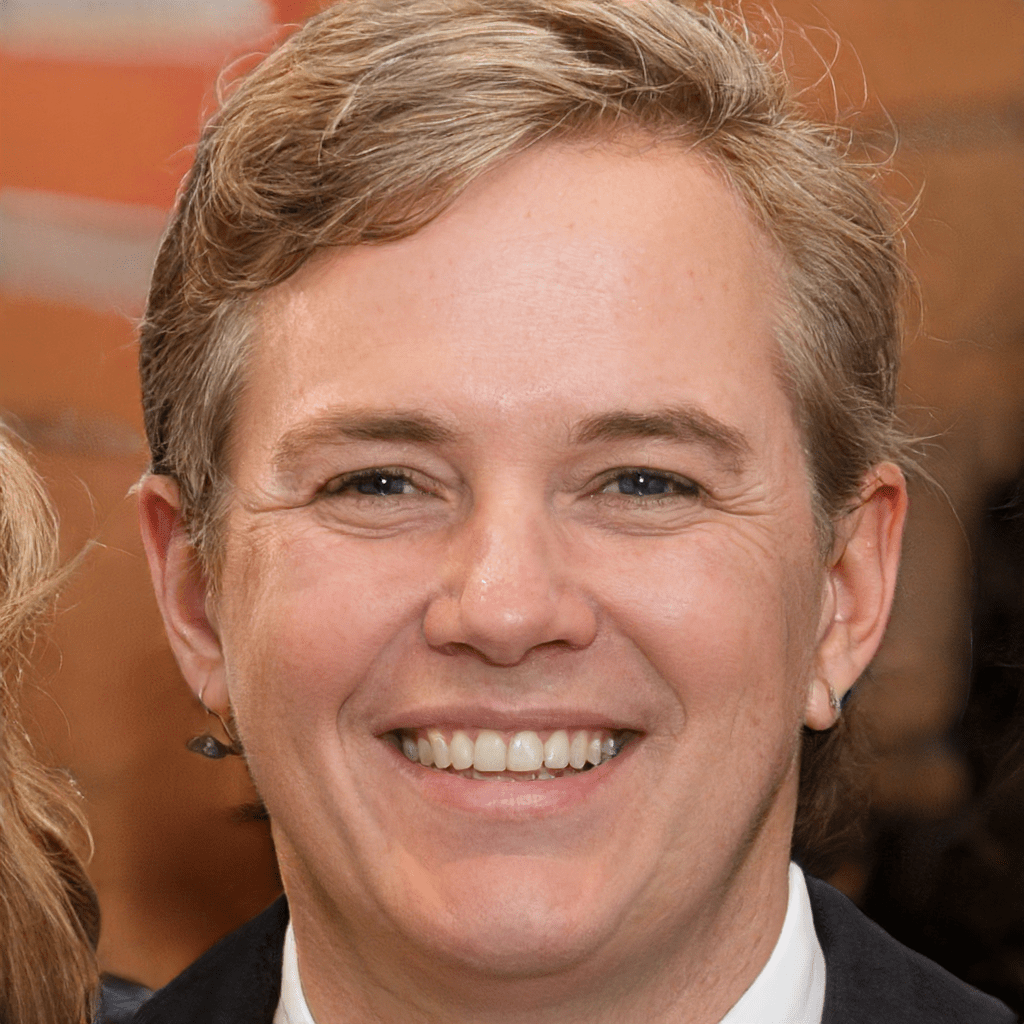} & 
 \includegraphics[width=0.12\linewidth]{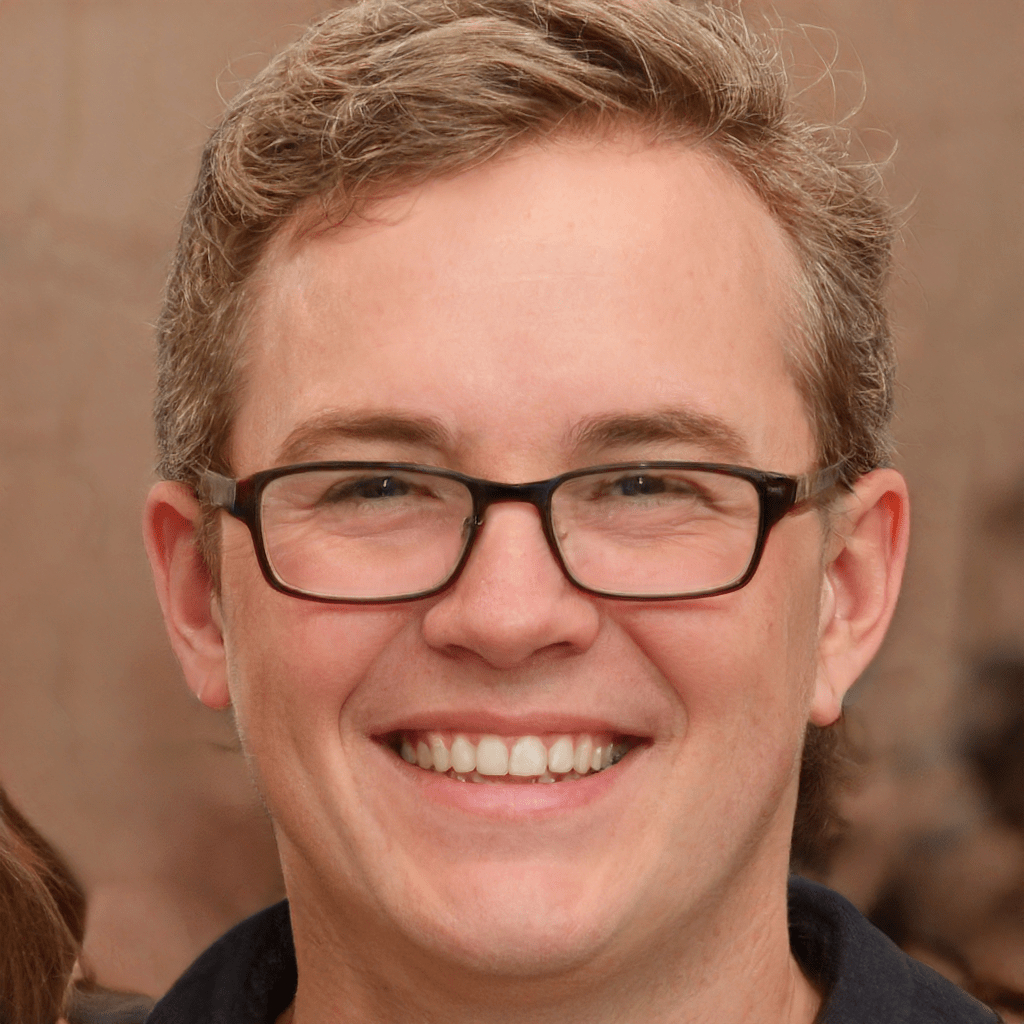} &
 \includegraphics[width=0.12\linewidth]{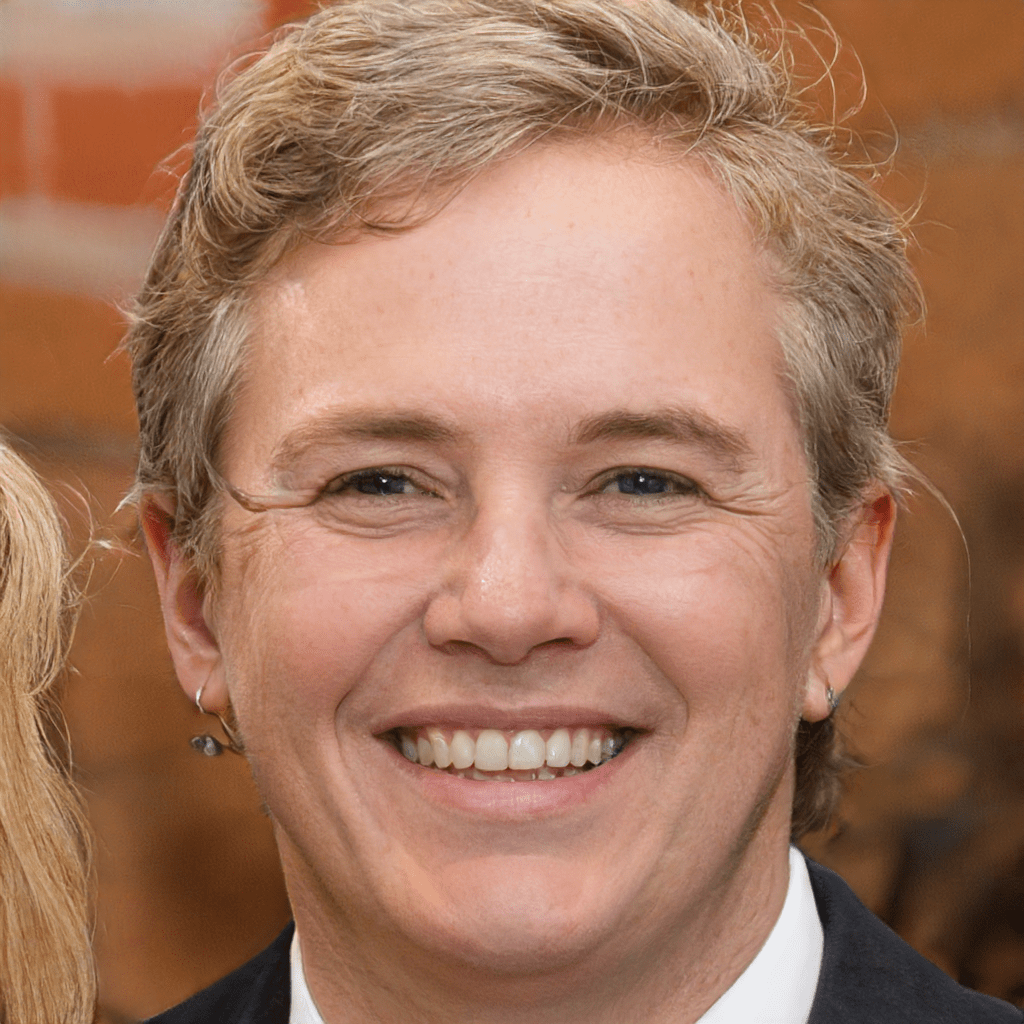} &
 \includegraphics[width=0.12\linewidth]{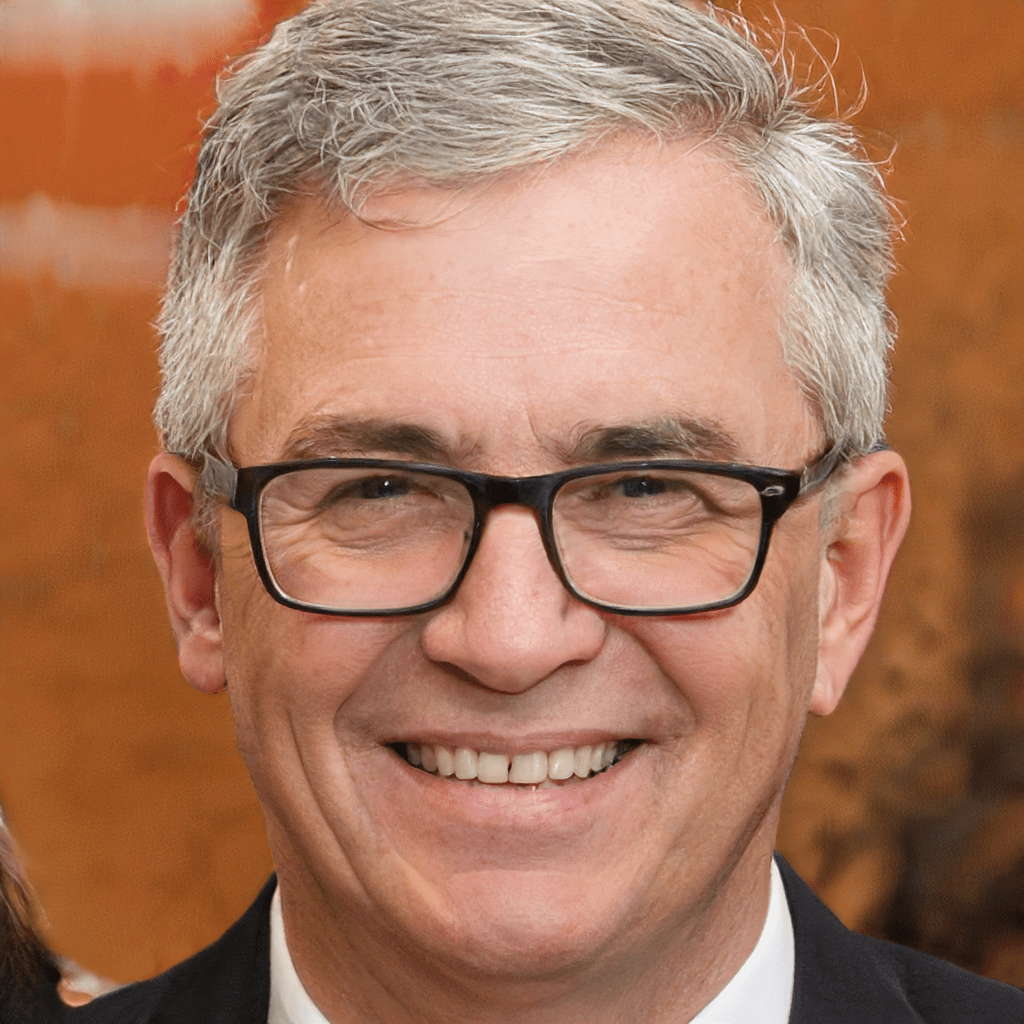} & 
 &
 \multirow{1}{*}[6ex]{\rotatebox[origin=c]{90}{\textbf{beard}}} &
 \includegraphics[width=0.12\linewidth]{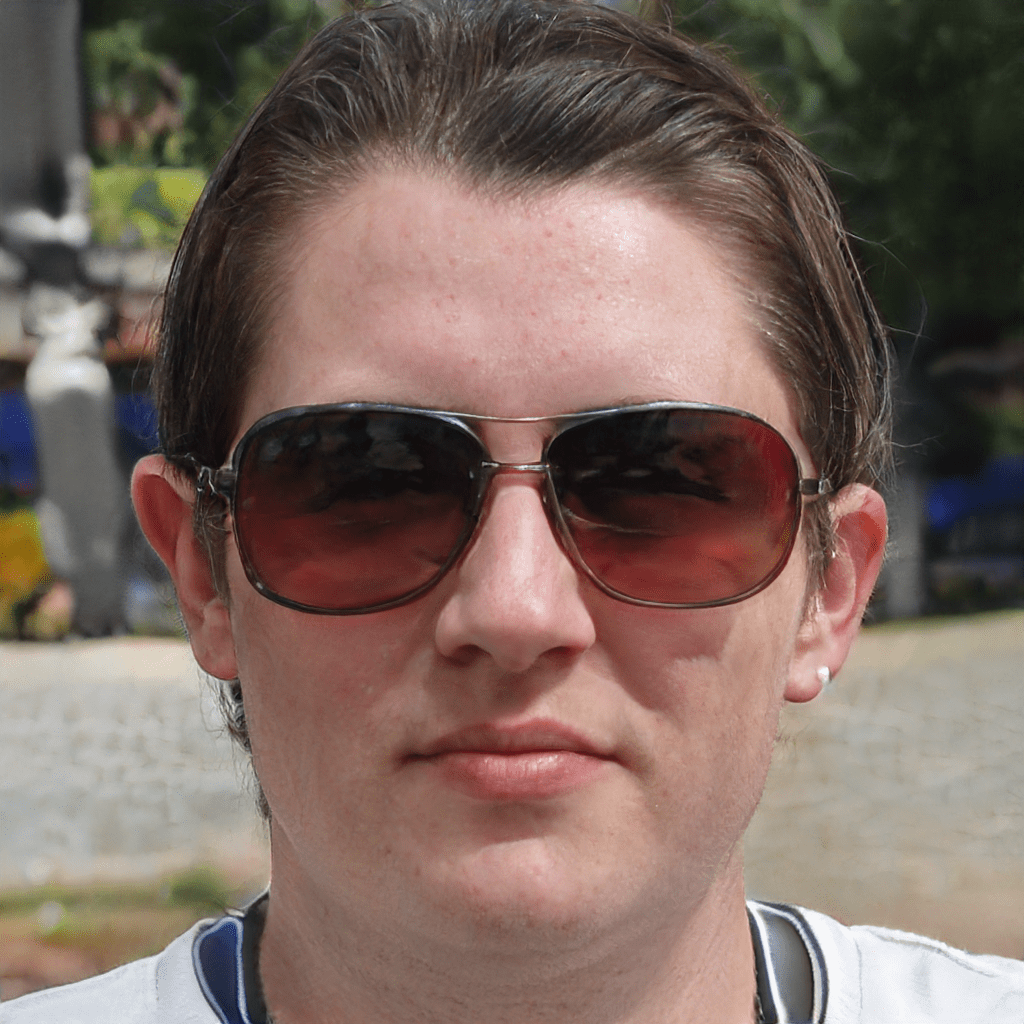}  &
 \includegraphics[width=0.12\linewidth]{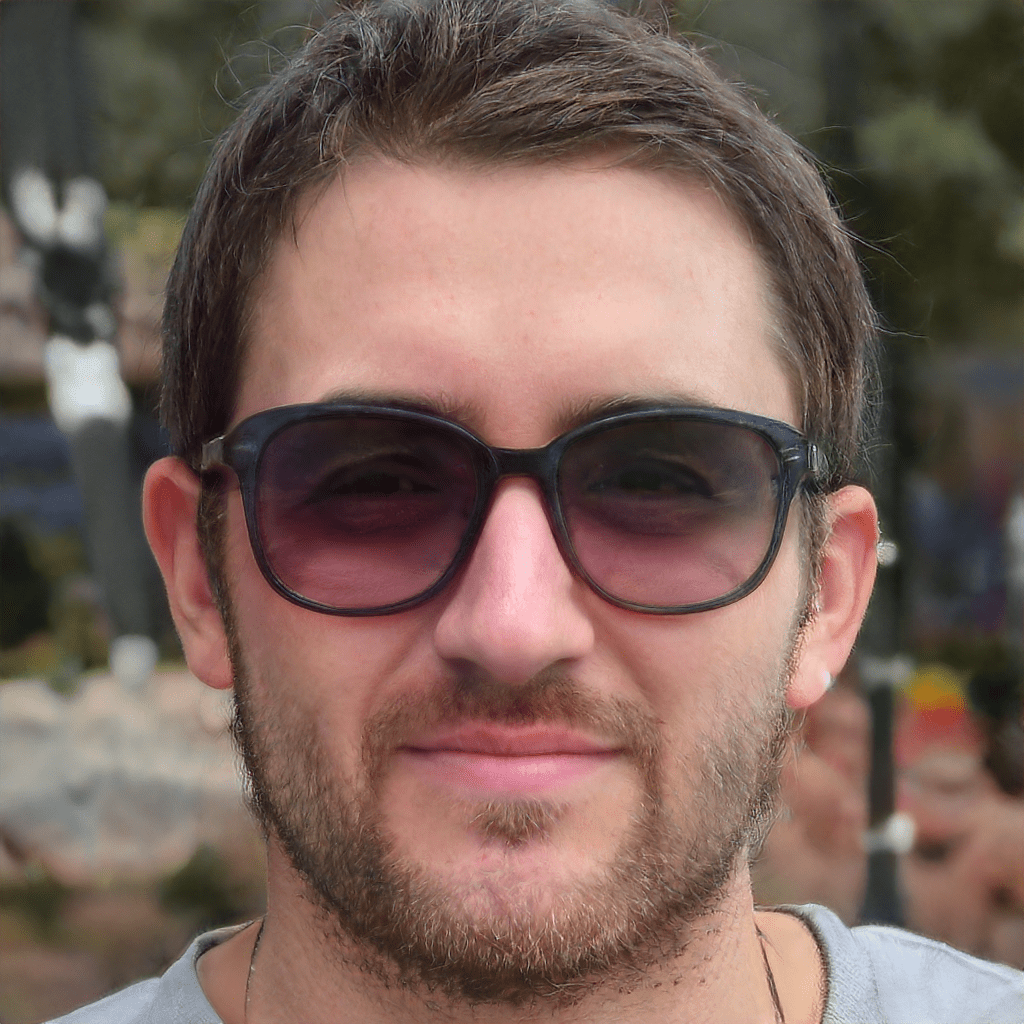} & 
 \includegraphics[width=0.12\linewidth]{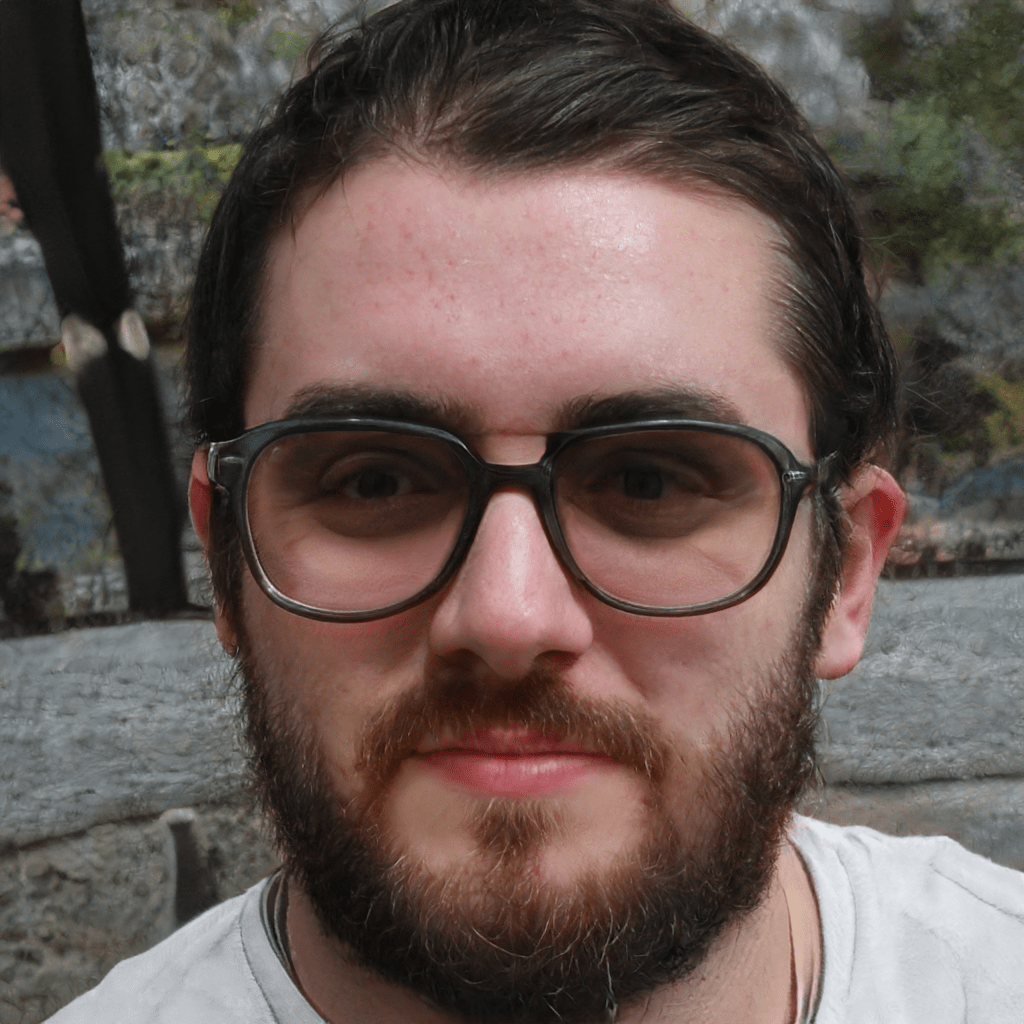} &
 \includegraphics[width=0.12\linewidth]{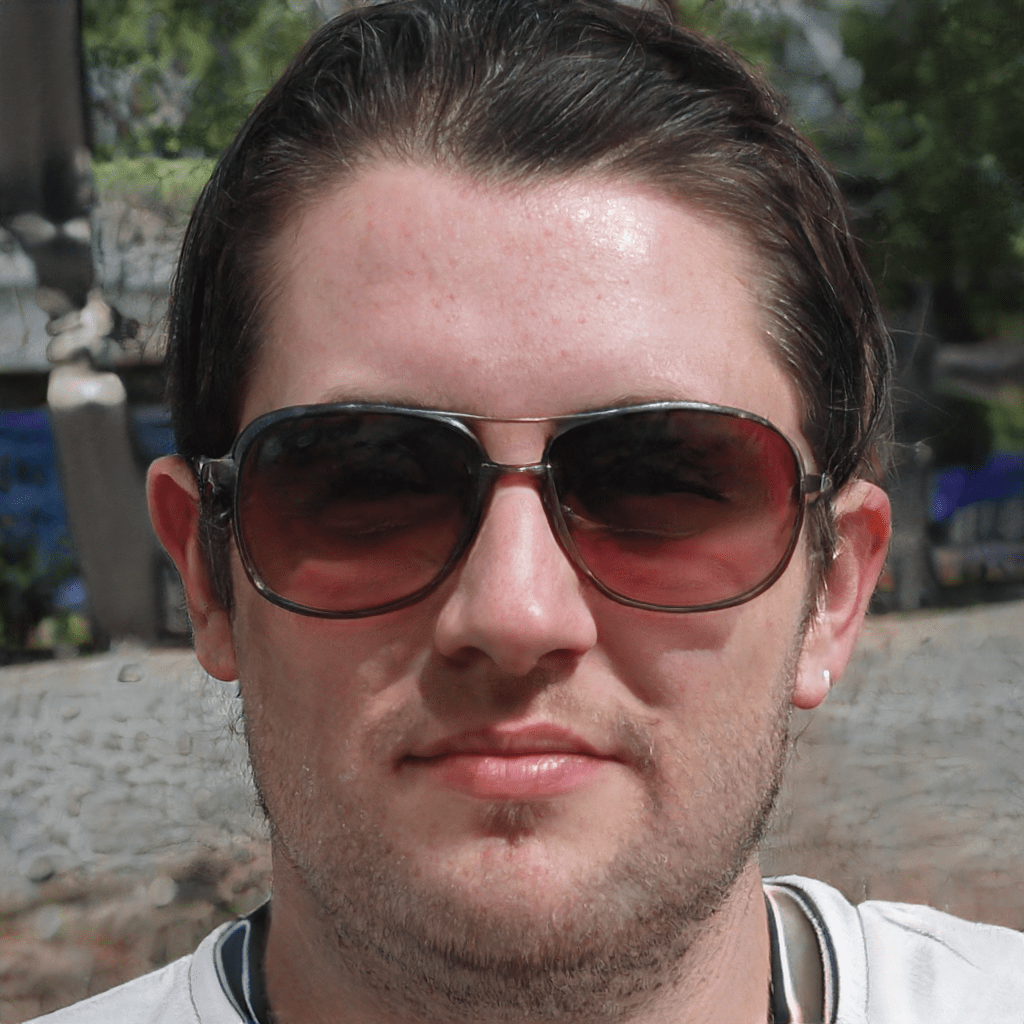}   \\
 \multirow{1}{*}[8.5ex]{\rotatebox[origin=c]{90}{\textbf{no smile}}} &
 \includegraphics[width=0.12\linewidth]{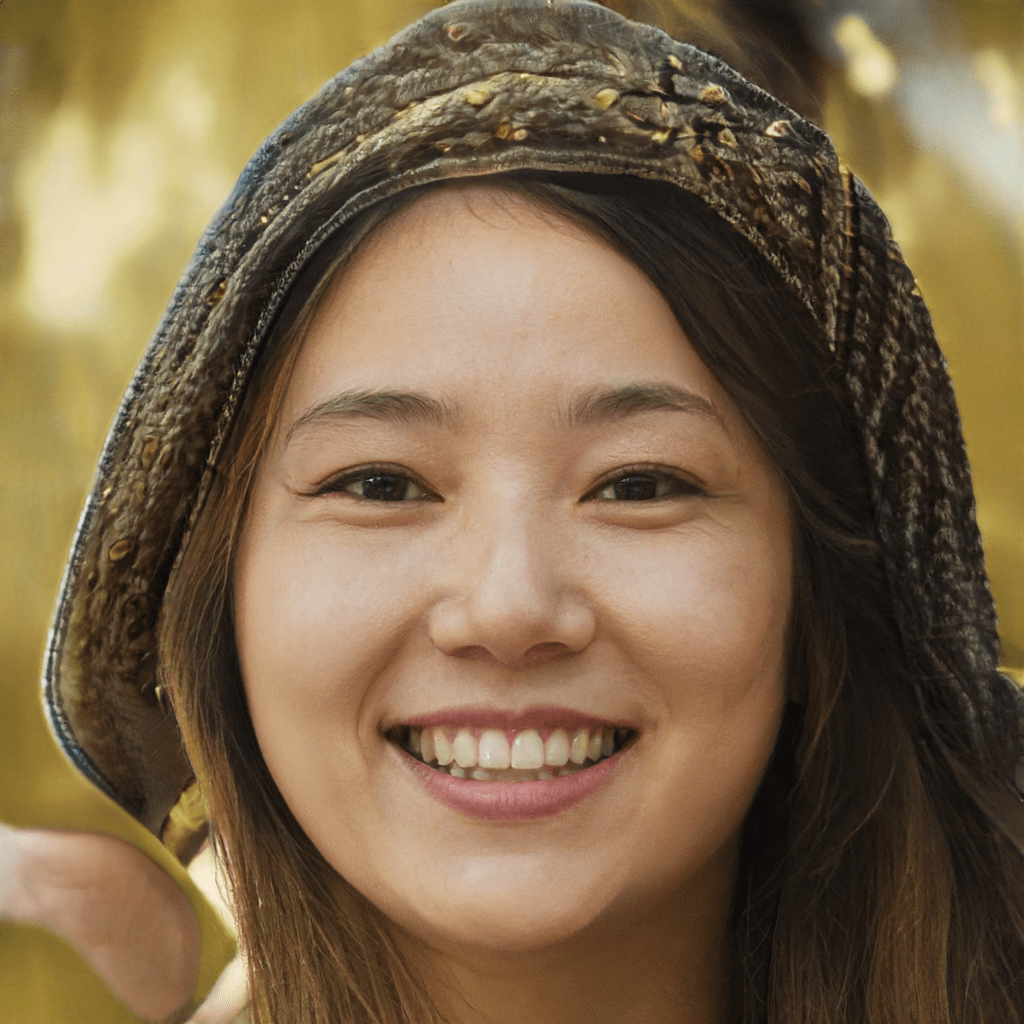} &
 \includegraphics[width=0.12\linewidth]{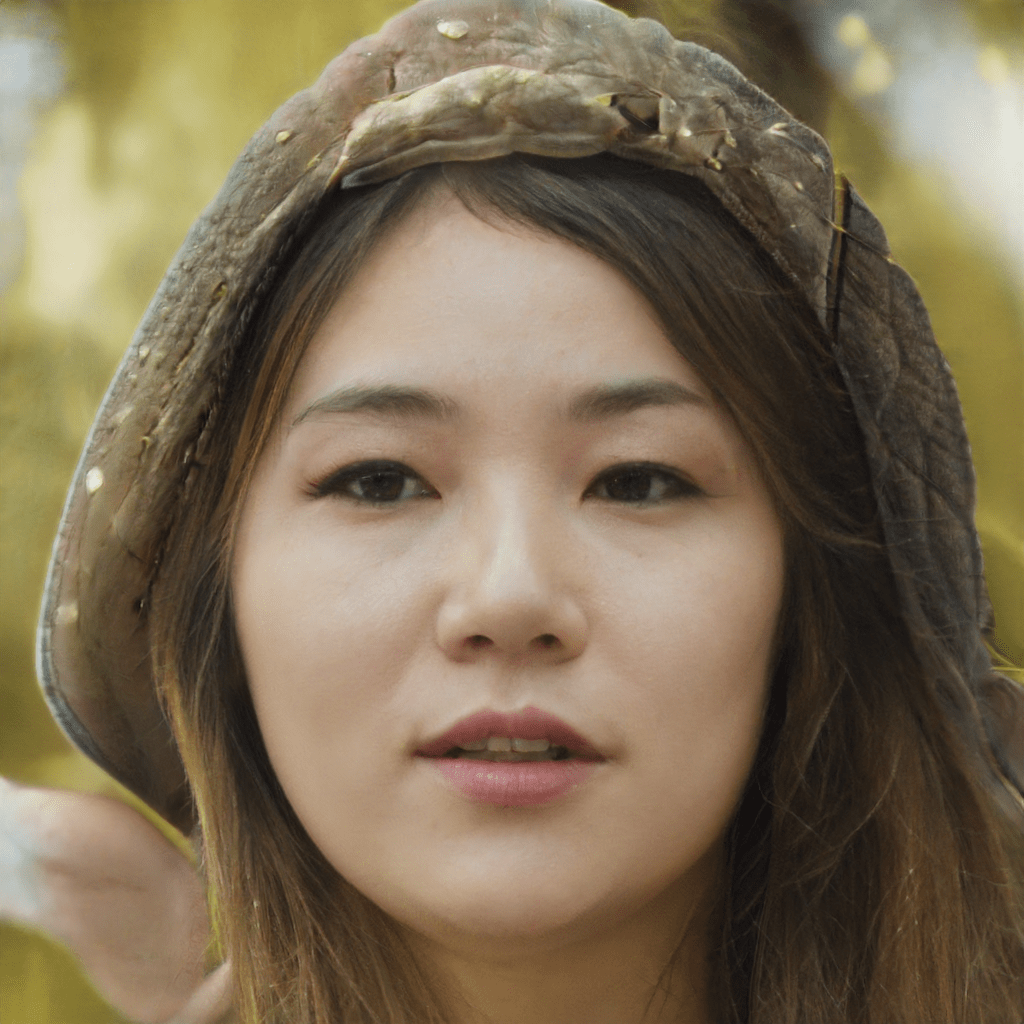} &
 \includegraphics[width=0.12\linewidth]{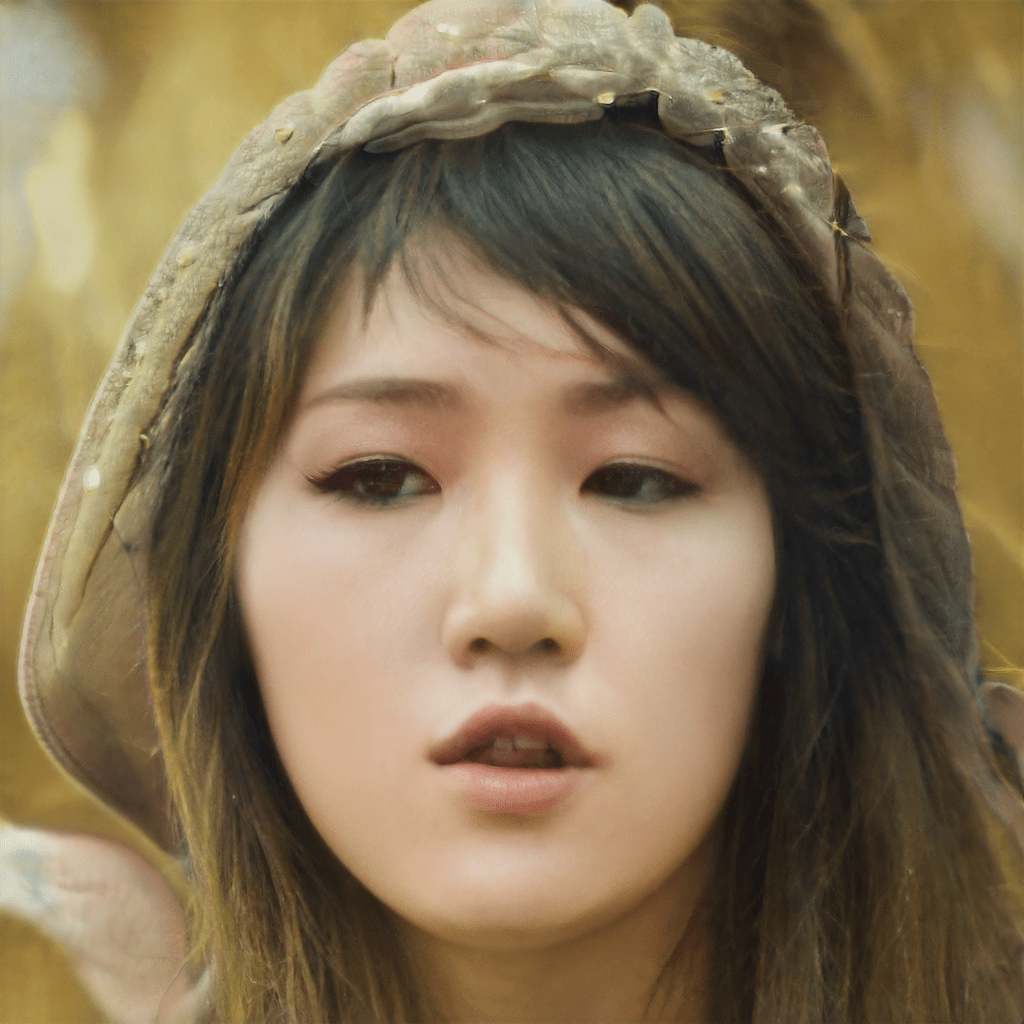} &
 \includegraphics[width=0.12\linewidth]{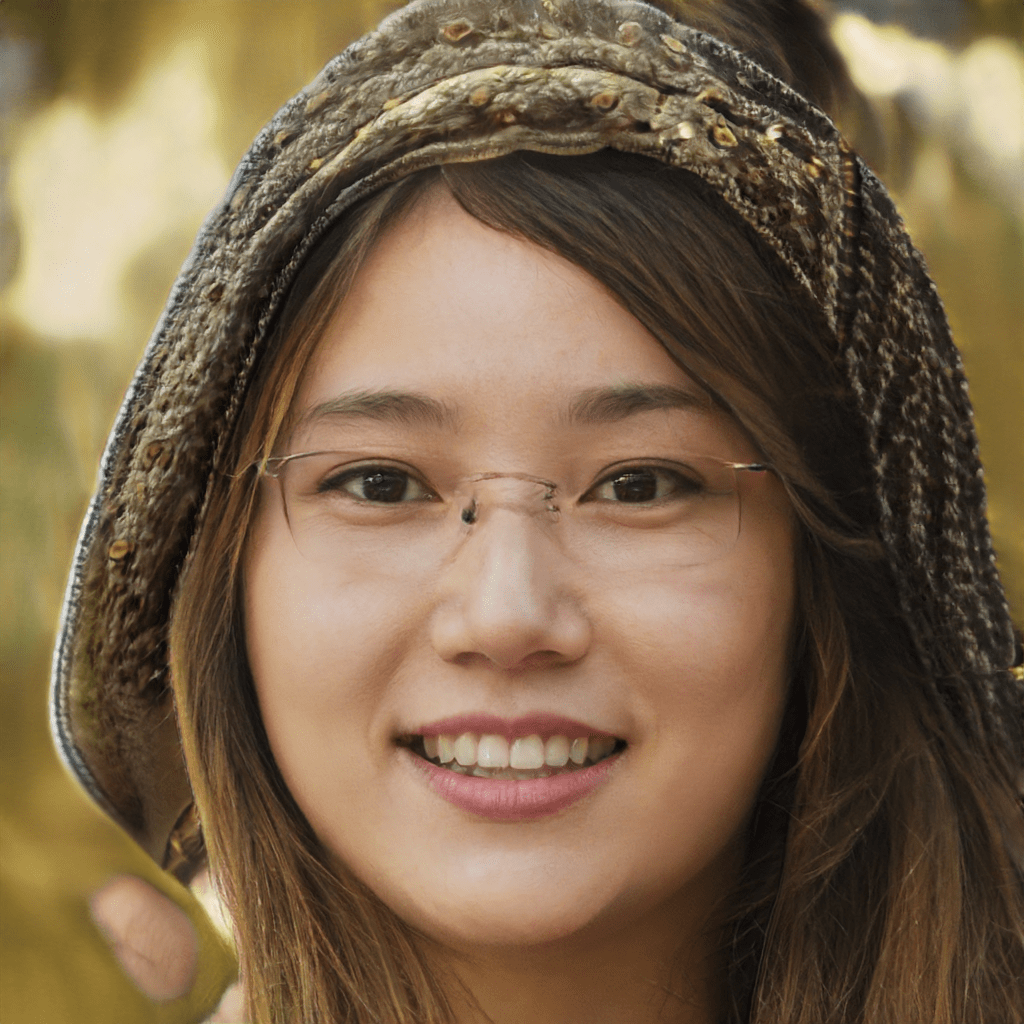} & 
 &
 \multirow{1}{*}[6ex]{\rotatebox[origin=c]{90}{\textbf{smile}}} &
 \includegraphics[width=0.12\linewidth]{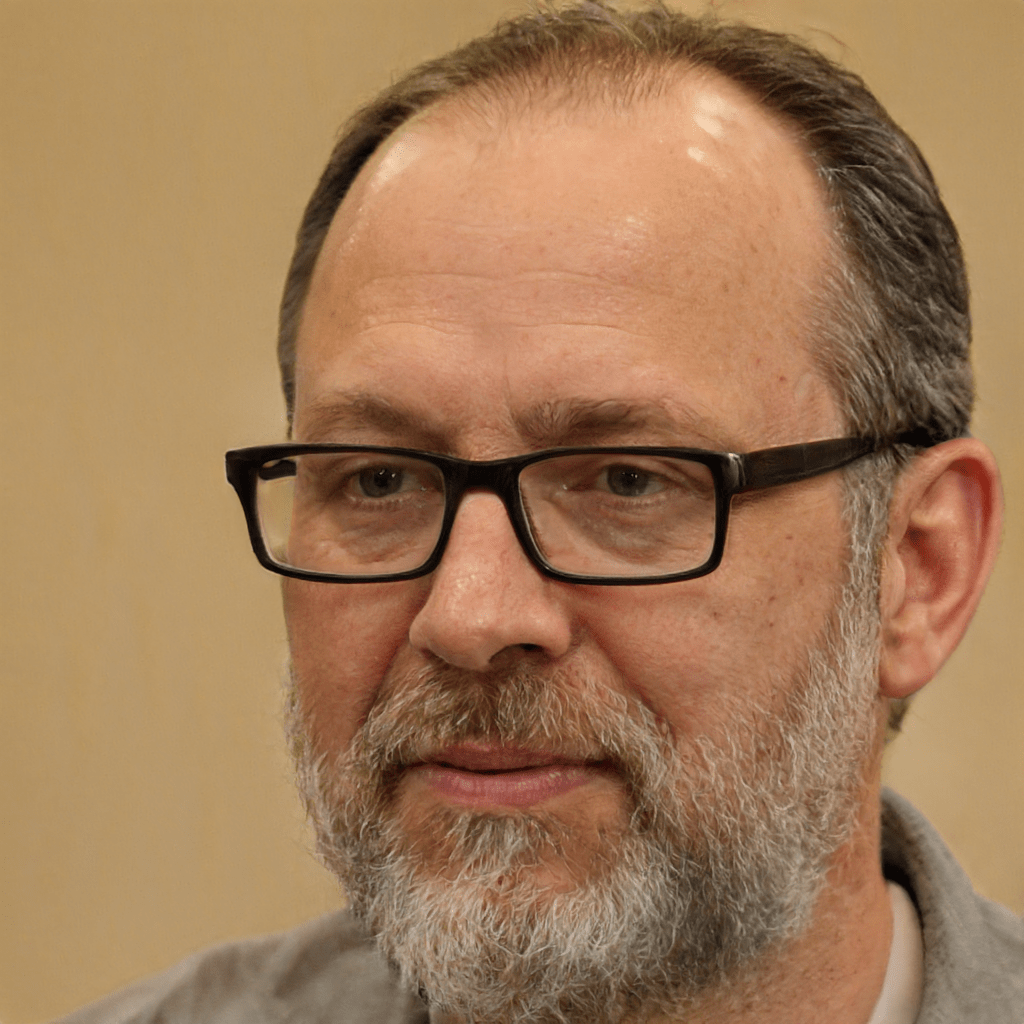} & 
 \includegraphics[width=0.12\linewidth]{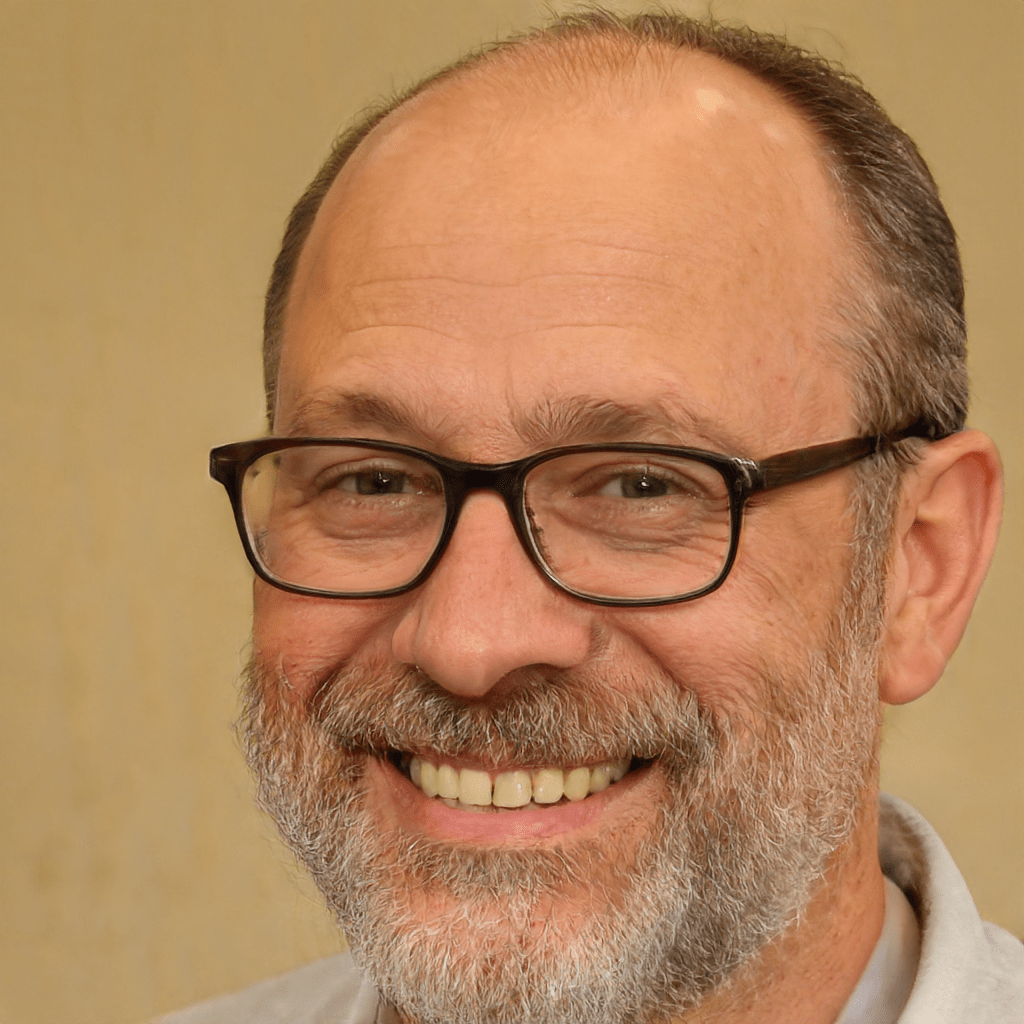} &
 \includegraphics[width=0.12\linewidth]{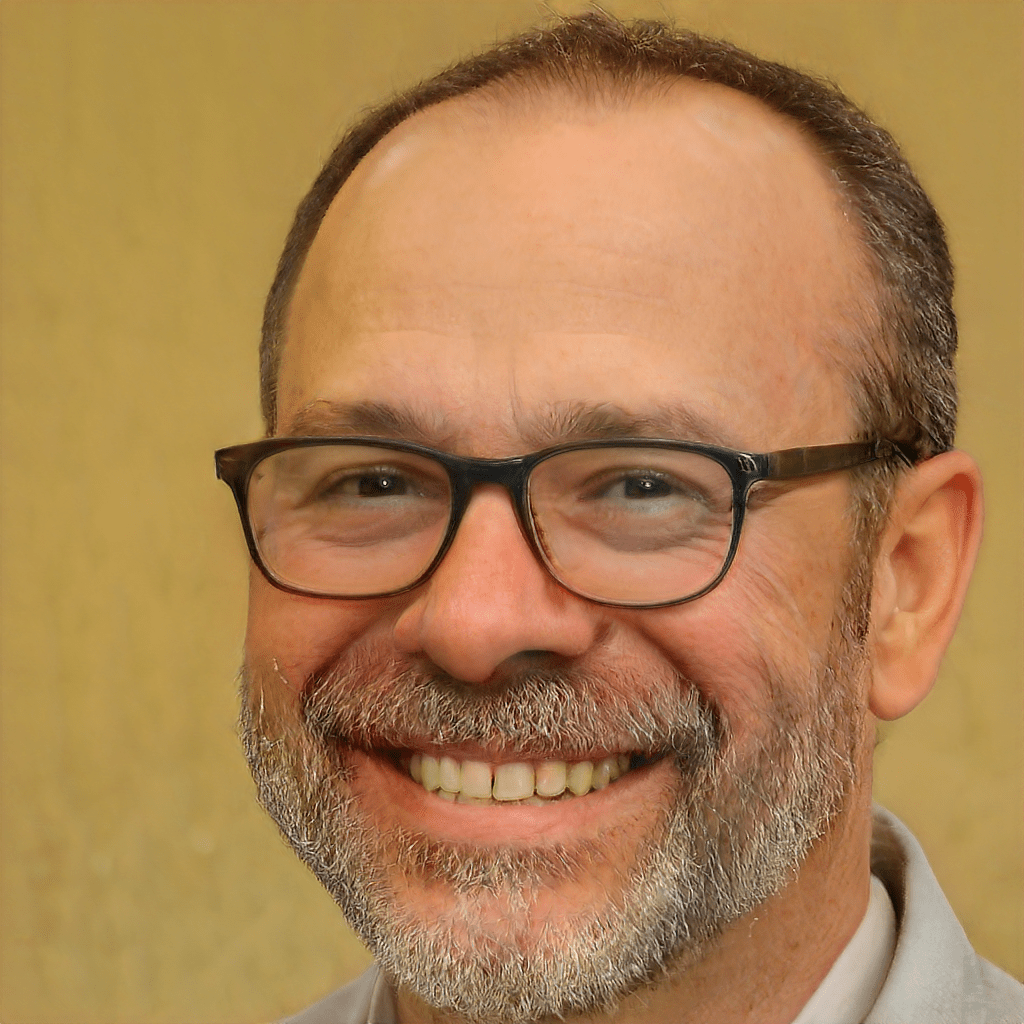} &
  \includegraphics[width=0.12\linewidth]{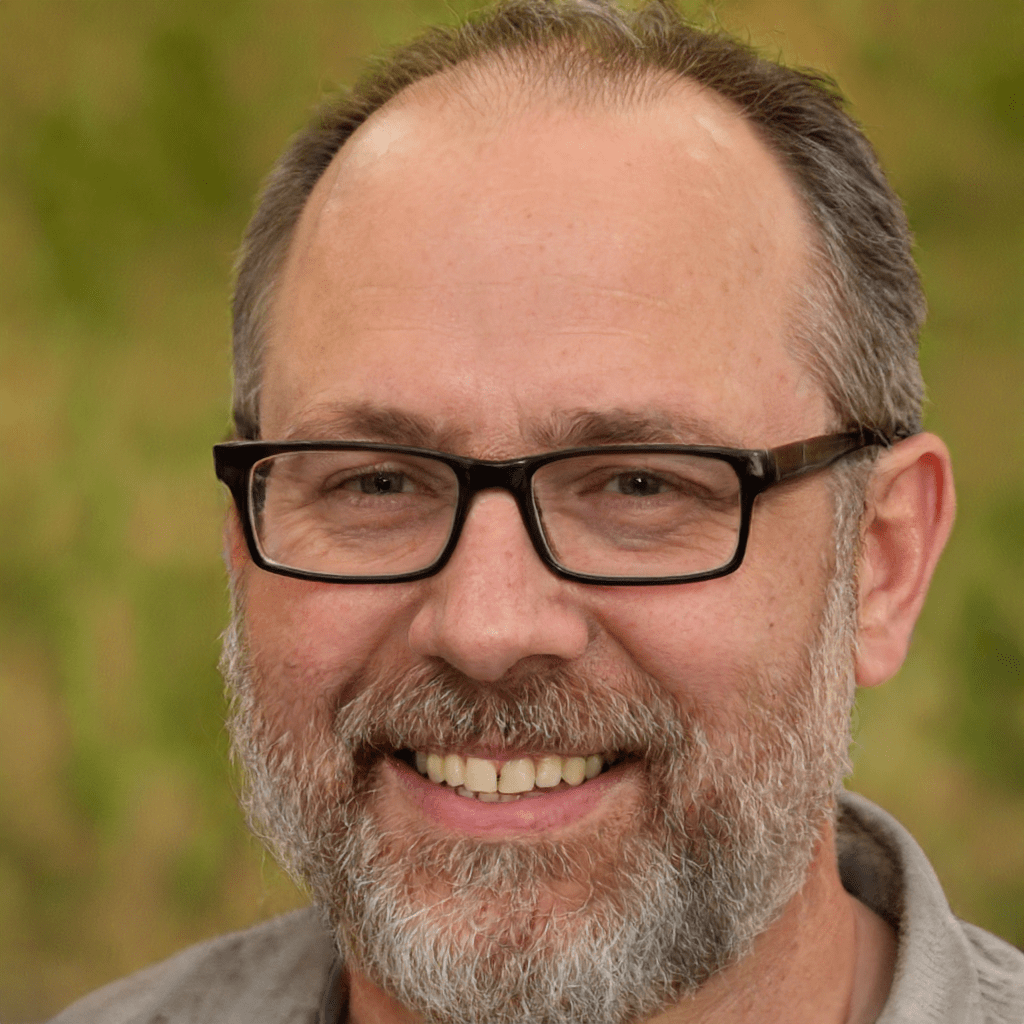}   
\end{tabular}
}
\caption{\textbf{Examples of attributes modification for all of the tested models on FFHQ dataset.} One can observe that \our{} correctly modifies the requested attributes and is less invasive to the remaining characteristics of the image than the competitive flow-based methods.}
\label{modification}
\vspace{-2em}
\end{figure}

\subsection{Discussion}

\paragraph{Image Editing:} Attribute manipulation in an image $x$ involves obtaining the style vector $w$. While a classification mechanism can assist by tagging generated images with desired attributes, this method was applied to our human facial features dataset, categorized externally using the Microsoft Face API.

However, for datasets lacking pre-tagged samples, a mechanism to retrieve $w$ is essential. Literature suggests various methods \cite{abdal2021clip2stylegan}, often using an iterative approach to approximate the StyleGAN latent vector $w$ for a given image $x$. This process, though, can be time-consuming and does not guarantee optimal results.

In our experiments, we used the method proposed in \cite{abdal2019image2stylegan} to retrieve $w$ for the AFHQv2 dataset, which did not initially come with the required latent vectors for the images.

\begin{figure}[]
\vspace{-1em}
\centering
\resizebox{\columnwidth}{!}{%
\begin{tabular}{ccccccc}
             &  & \textbf{glasses} & \textbf{beard} & \textbf{expression} & \textbf{young} & \textbf{old} \\
\multirow{3}{*}[-3ex]{\includegraphics[width=0.16\linewidth]{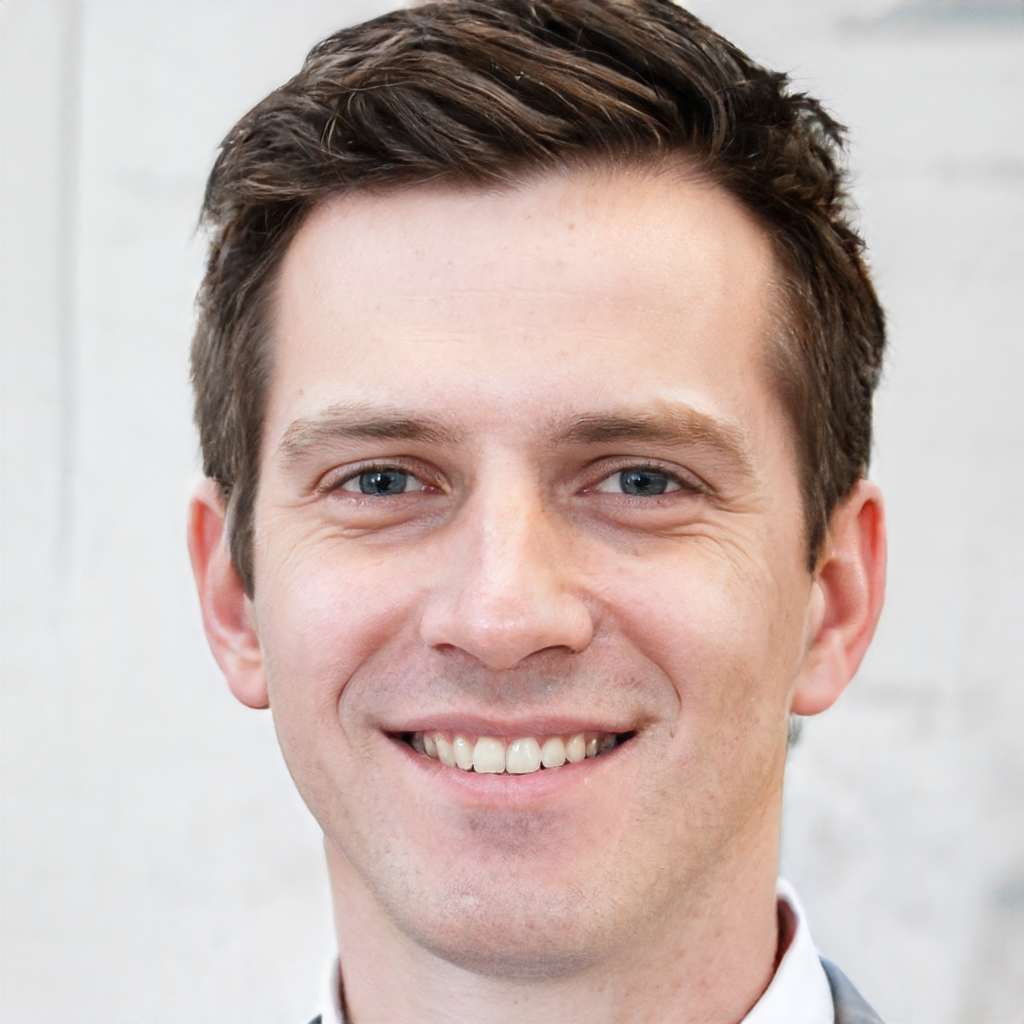}} &
\multirow{1}{*}[+10ex]{\rotatebox[]{90}{\small{\textbf{StyleAE}}}} &
\includegraphics[width=0.16\linewidth]{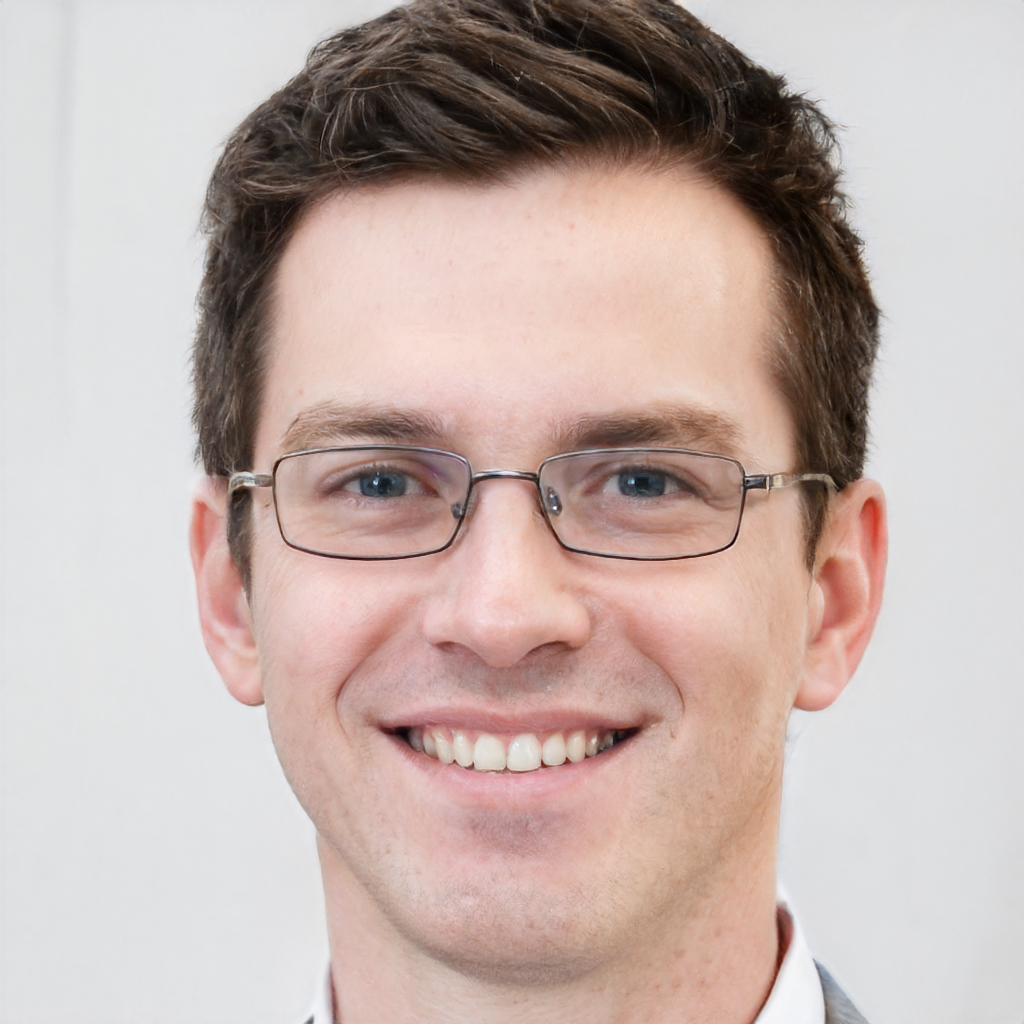} & 
\includegraphics[width=0.16\linewidth]{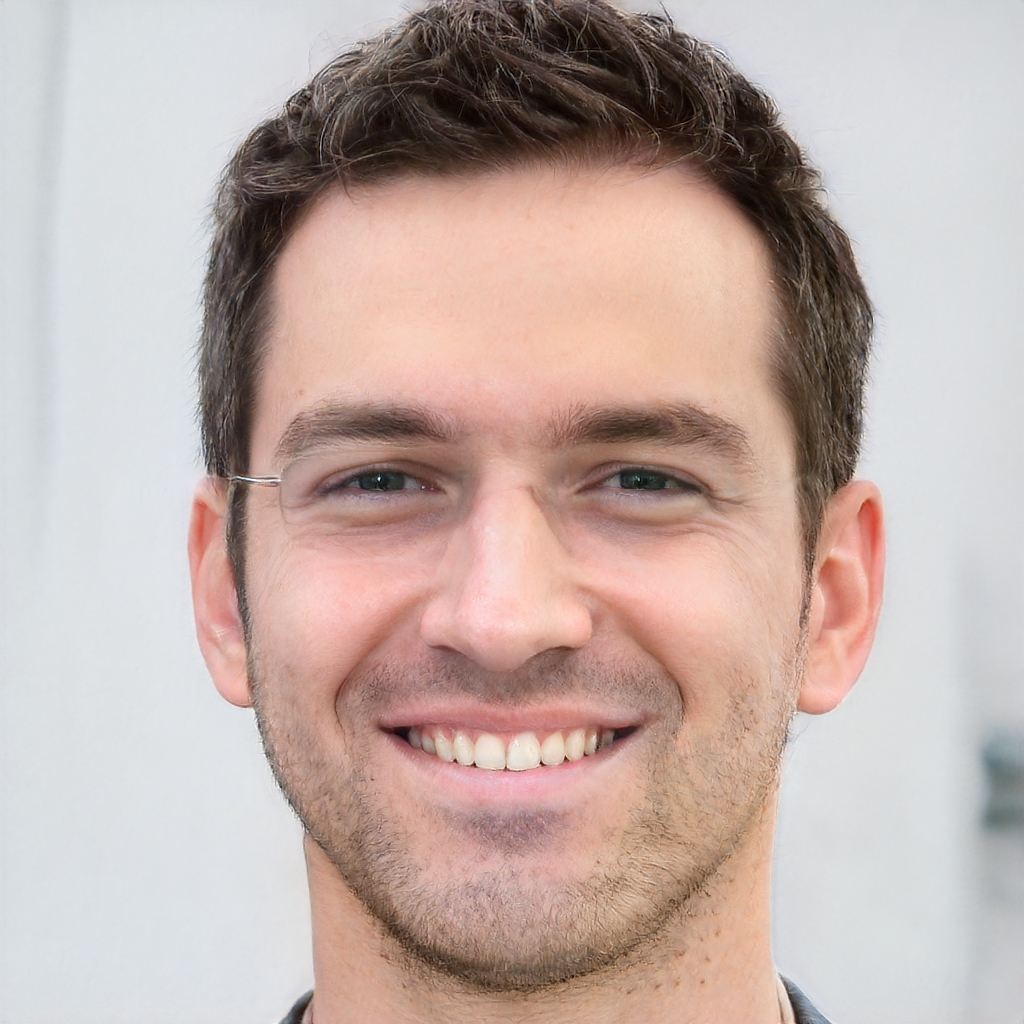} &
\includegraphics[width=0.16\linewidth]{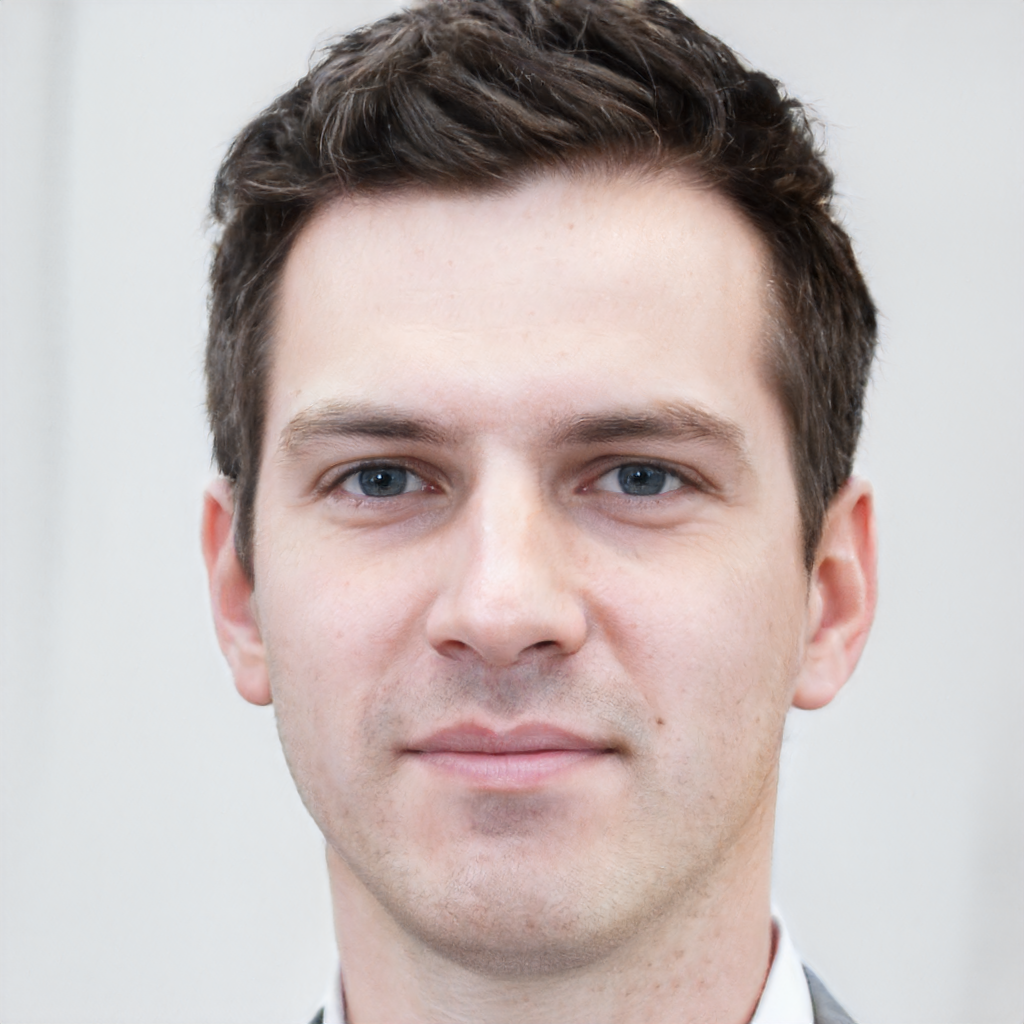} & 
\includegraphics[width=0.16\linewidth]{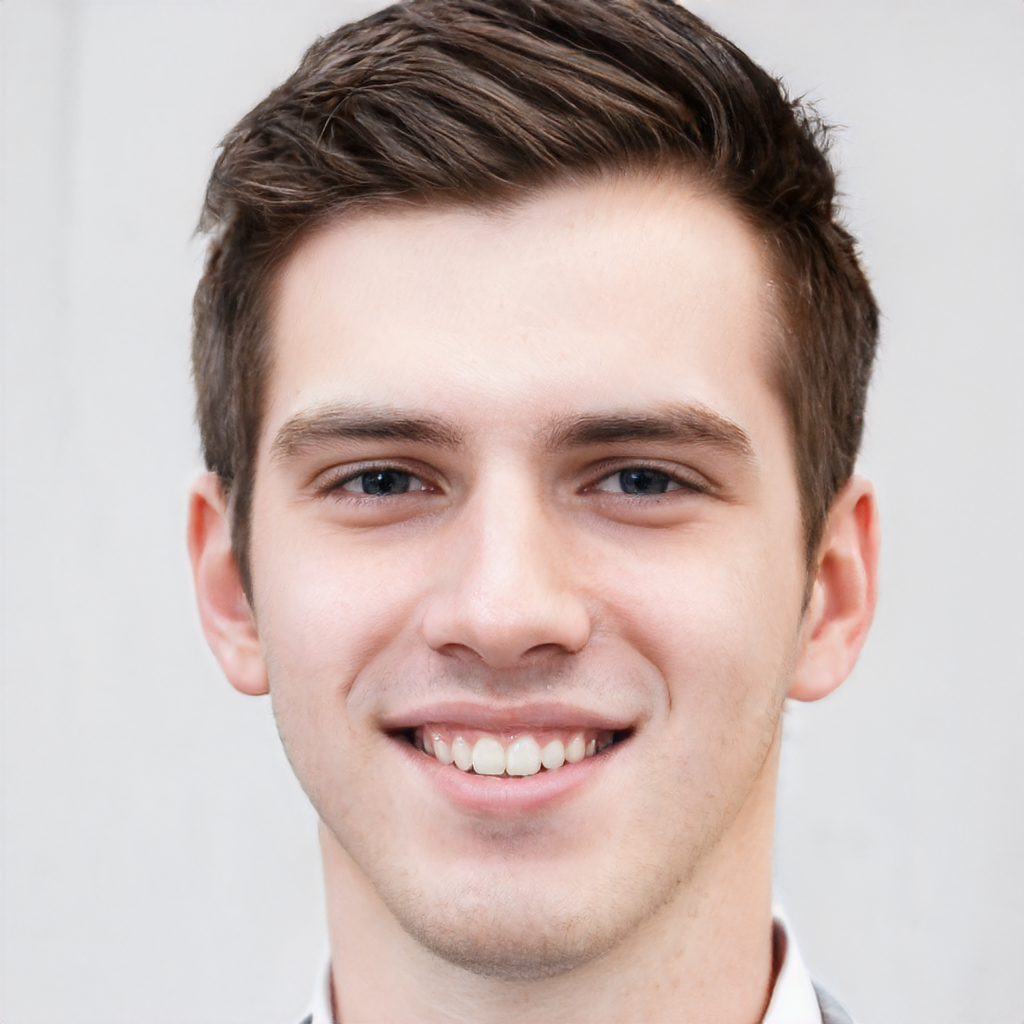} &
\includegraphics[width=0.16\linewidth]{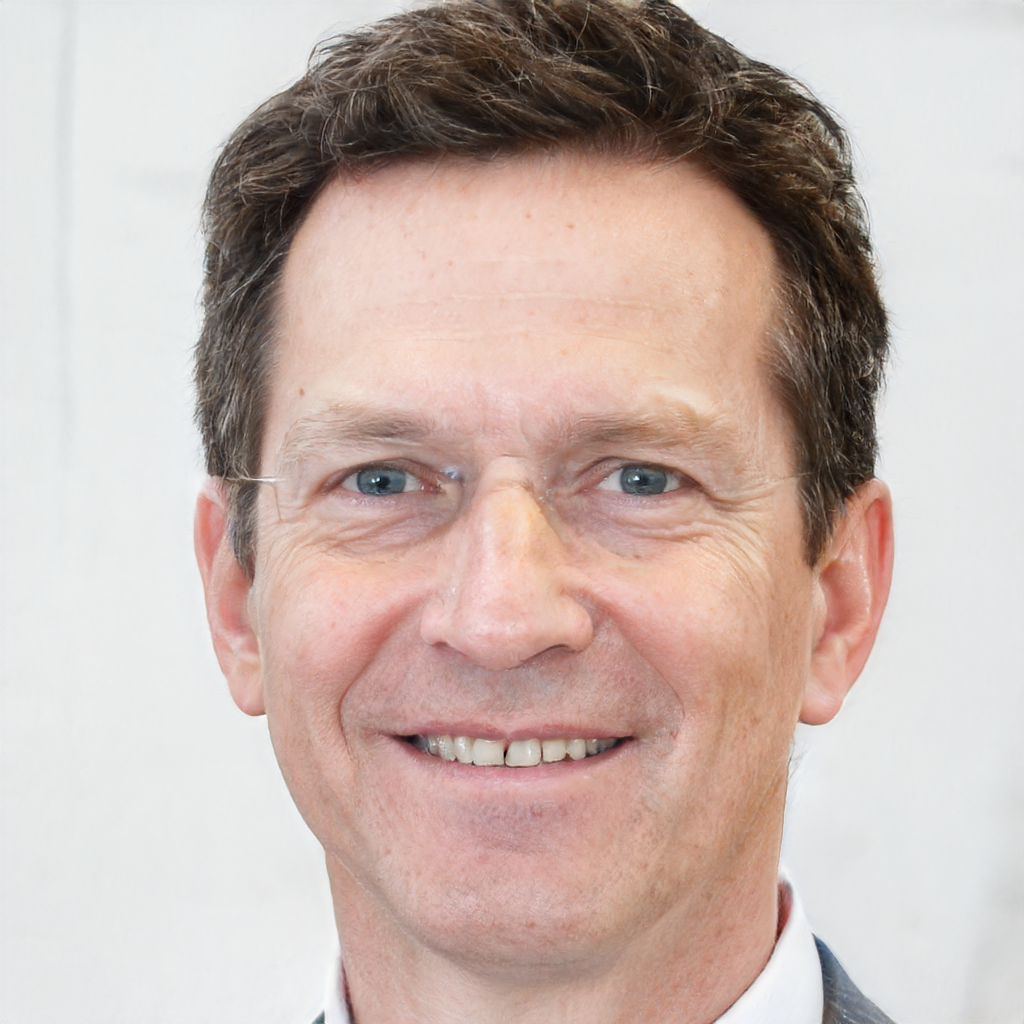} \\
                  &
\multirow{1}{*}[+11ex]{\rotatebox[]{90}{\small{\textbf{StyleFlow}}}} &
\includegraphics[width=0.16\linewidth]{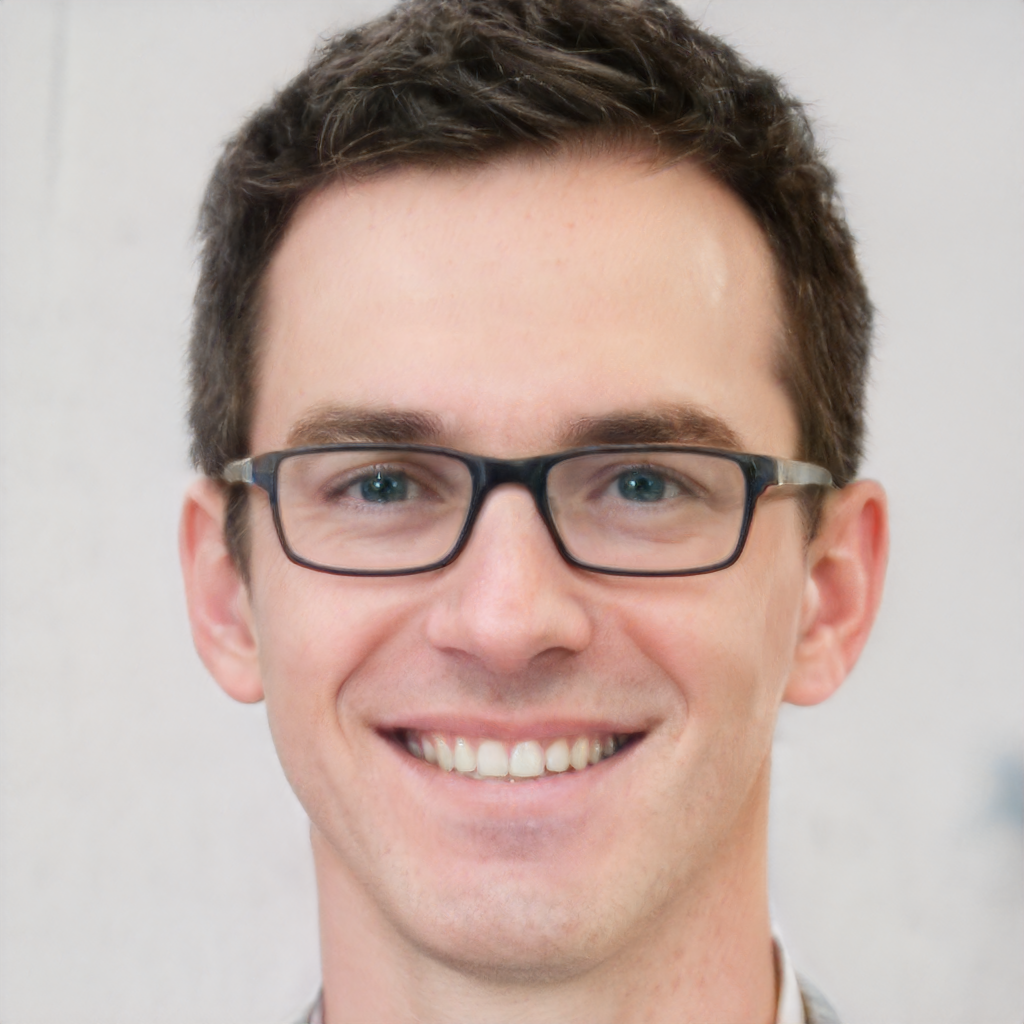} &  
\includegraphics[width=0.16\linewidth]{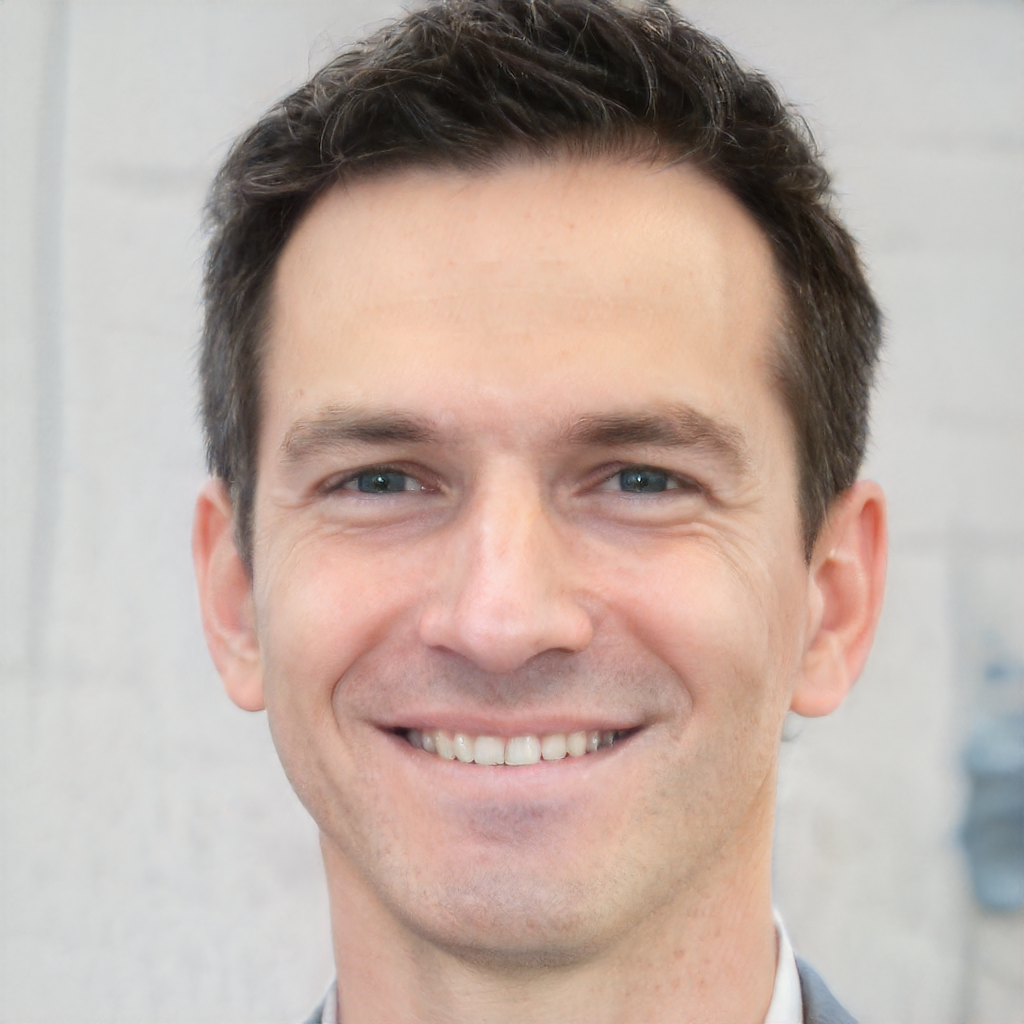}  & 
\includegraphics[width=0.16\linewidth]{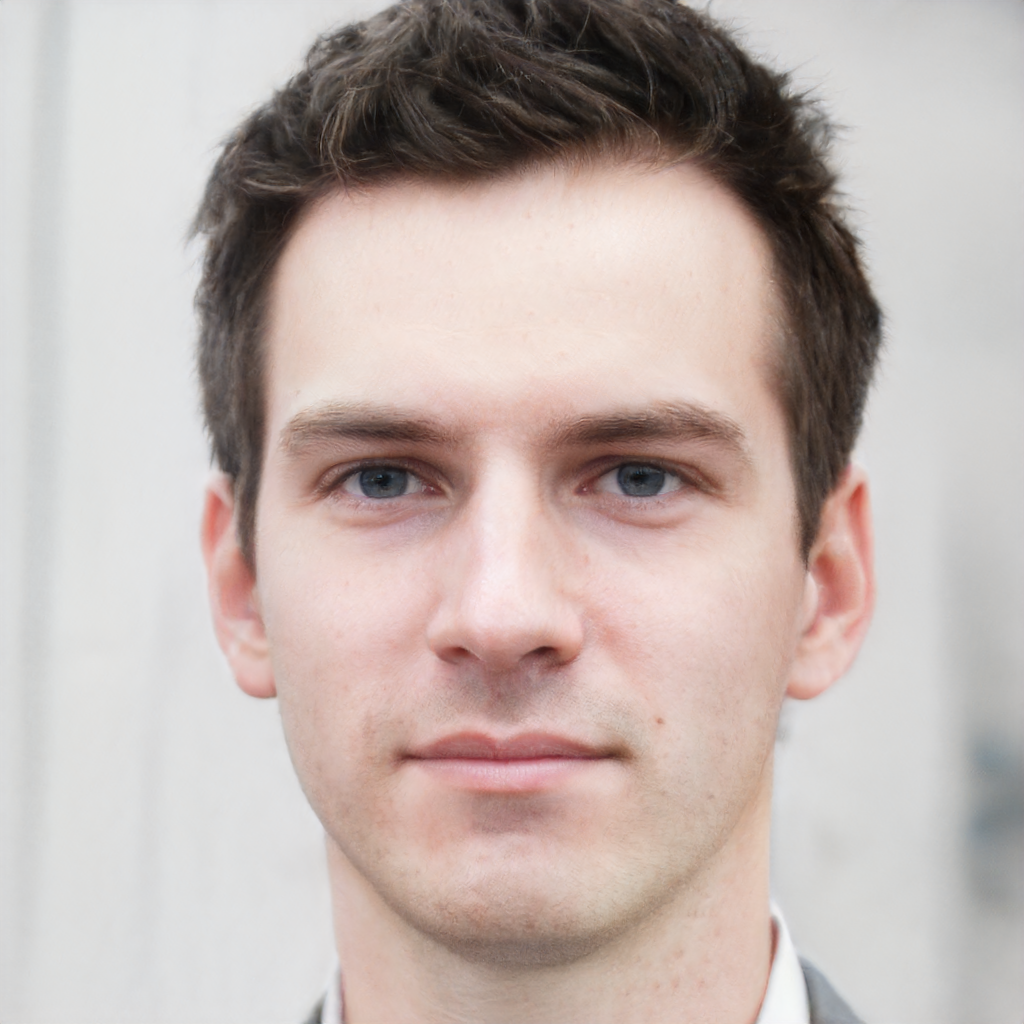} & 
\includegraphics[width=0.16\linewidth]{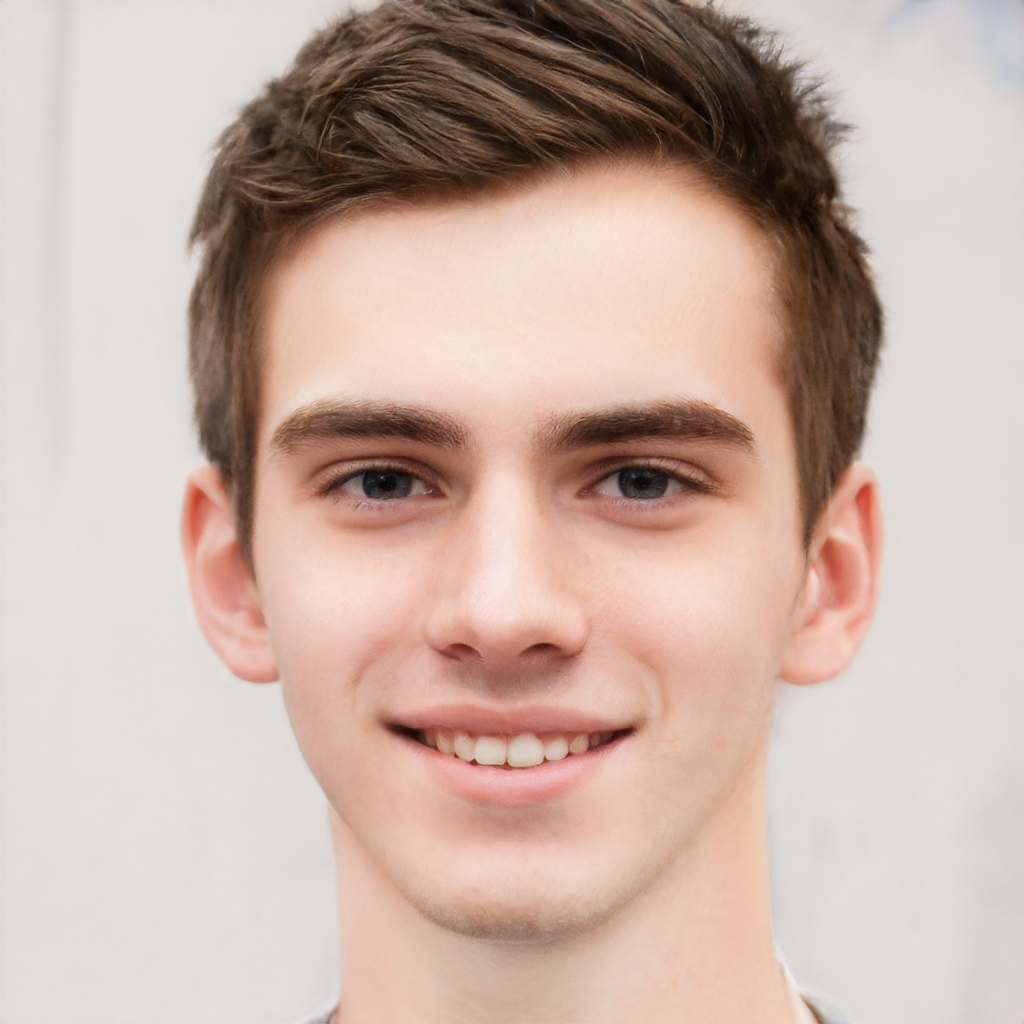} & 
\includegraphics[width=0.16\linewidth]{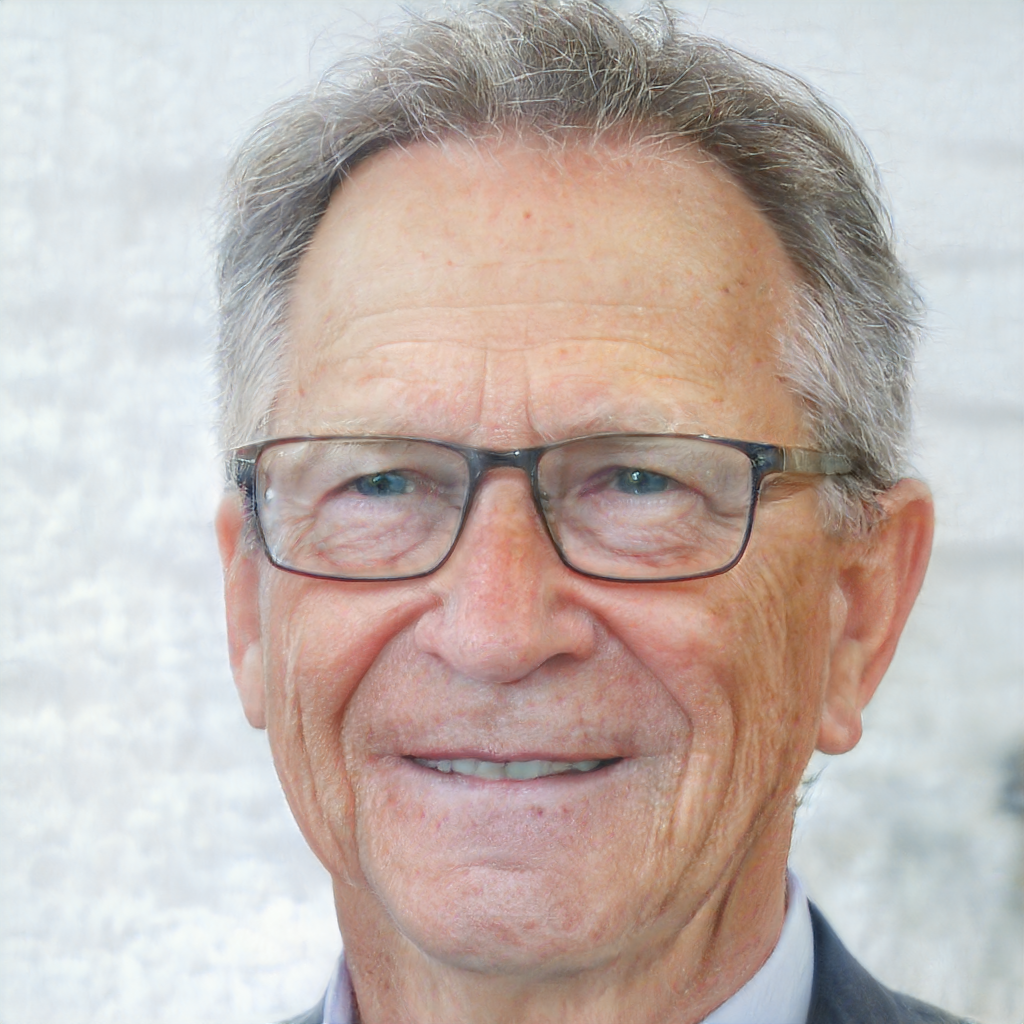}  \\
                  & 
\multirow{1}{*}[+10ex]{\rotatebox[]{90}{\small{\textbf{PluGeN}}}} &
\includegraphics[width=0.16\linewidth]{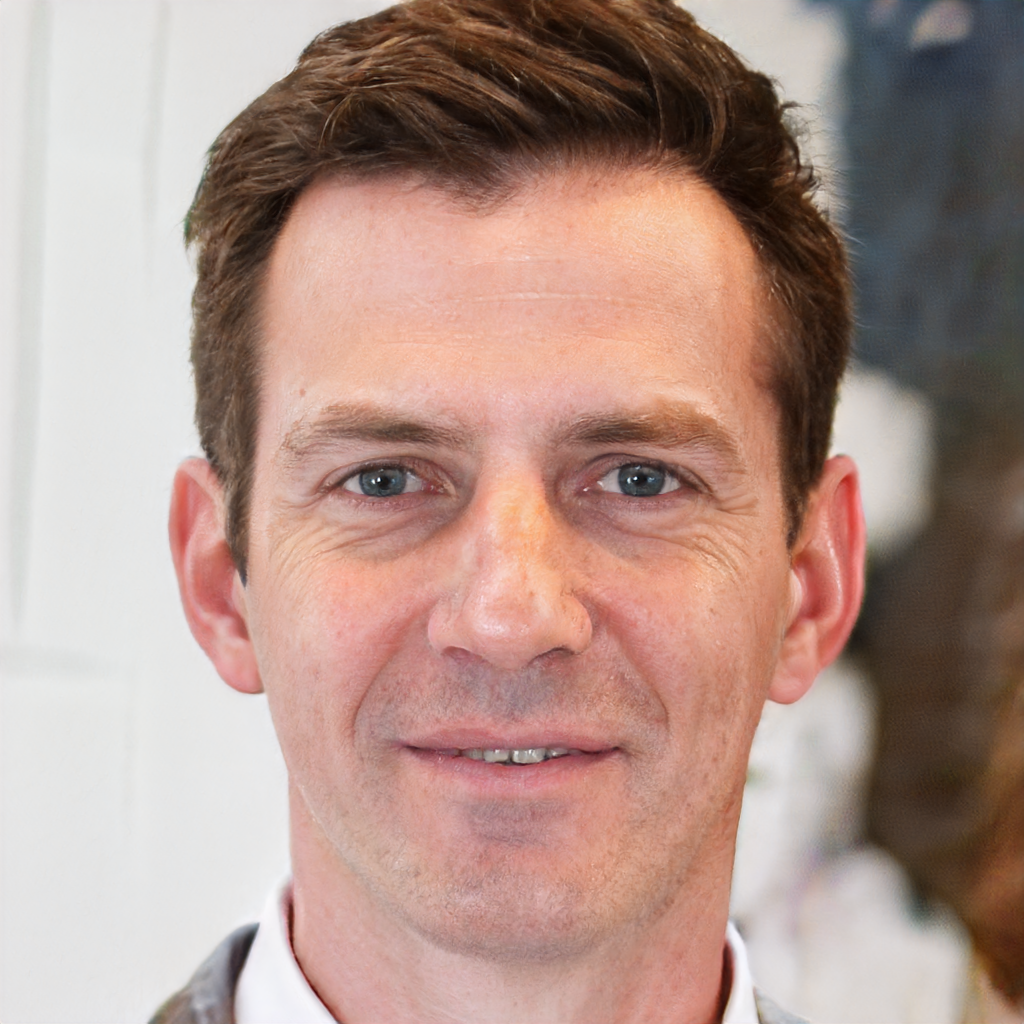} &
\includegraphics[width=0.16\linewidth]{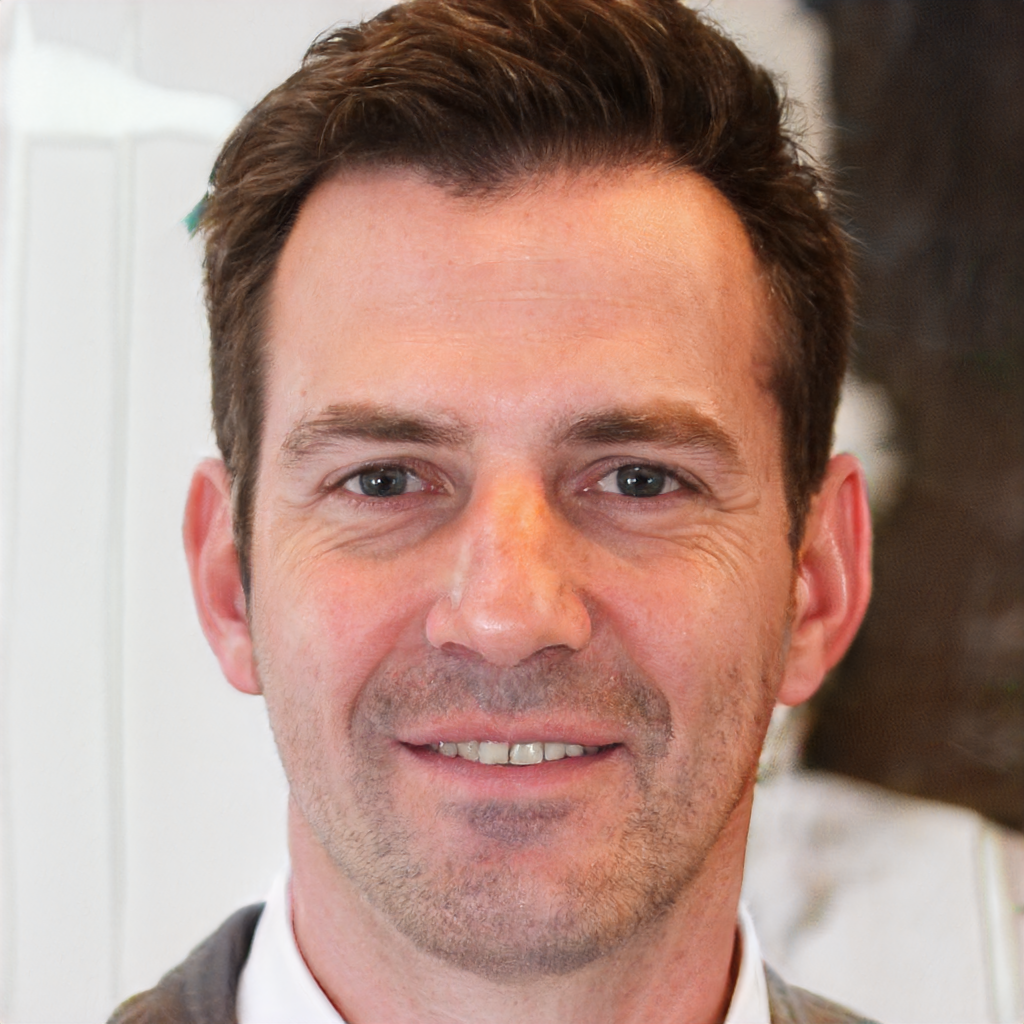} &
\includegraphics[width=0.16\linewidth]{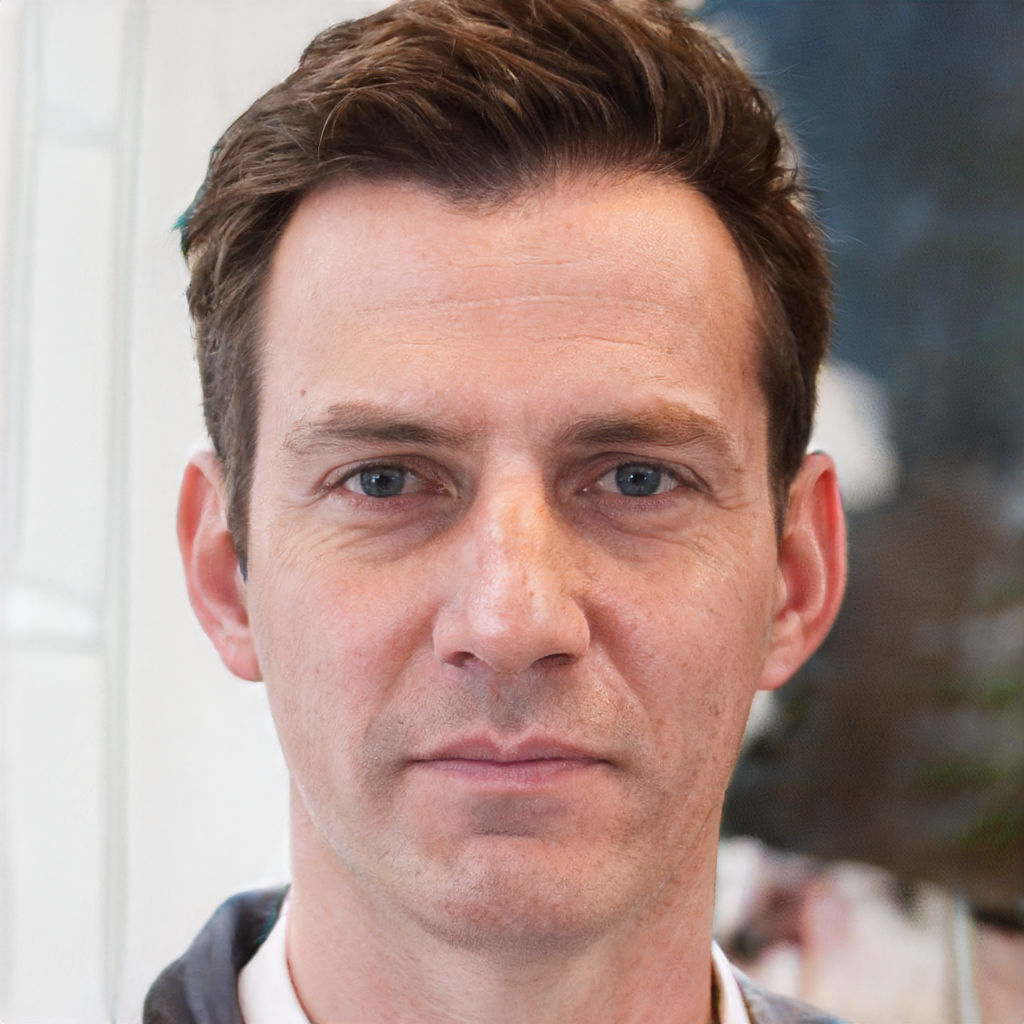} & 
\includegraphics[width=0.16\linewidth]{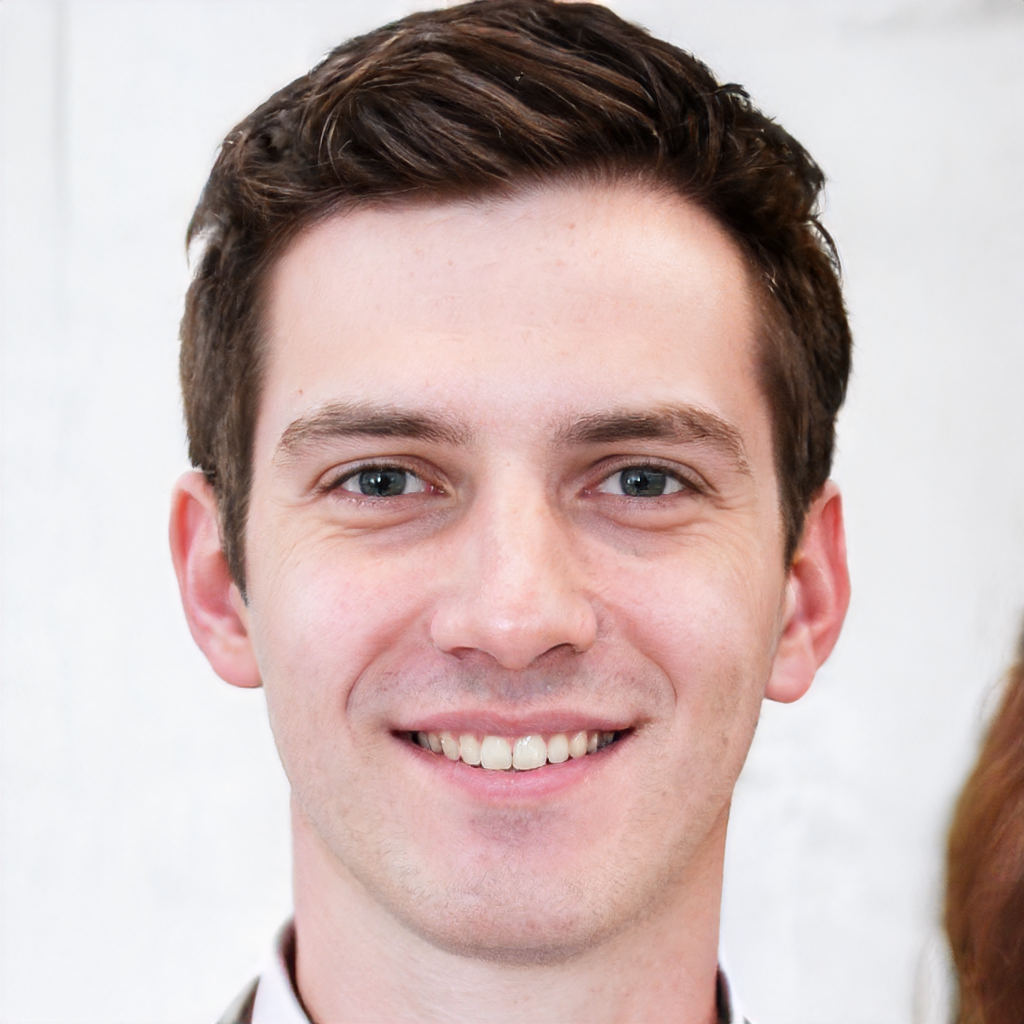} & 
\includegraphics[width=0.16\linewidth]{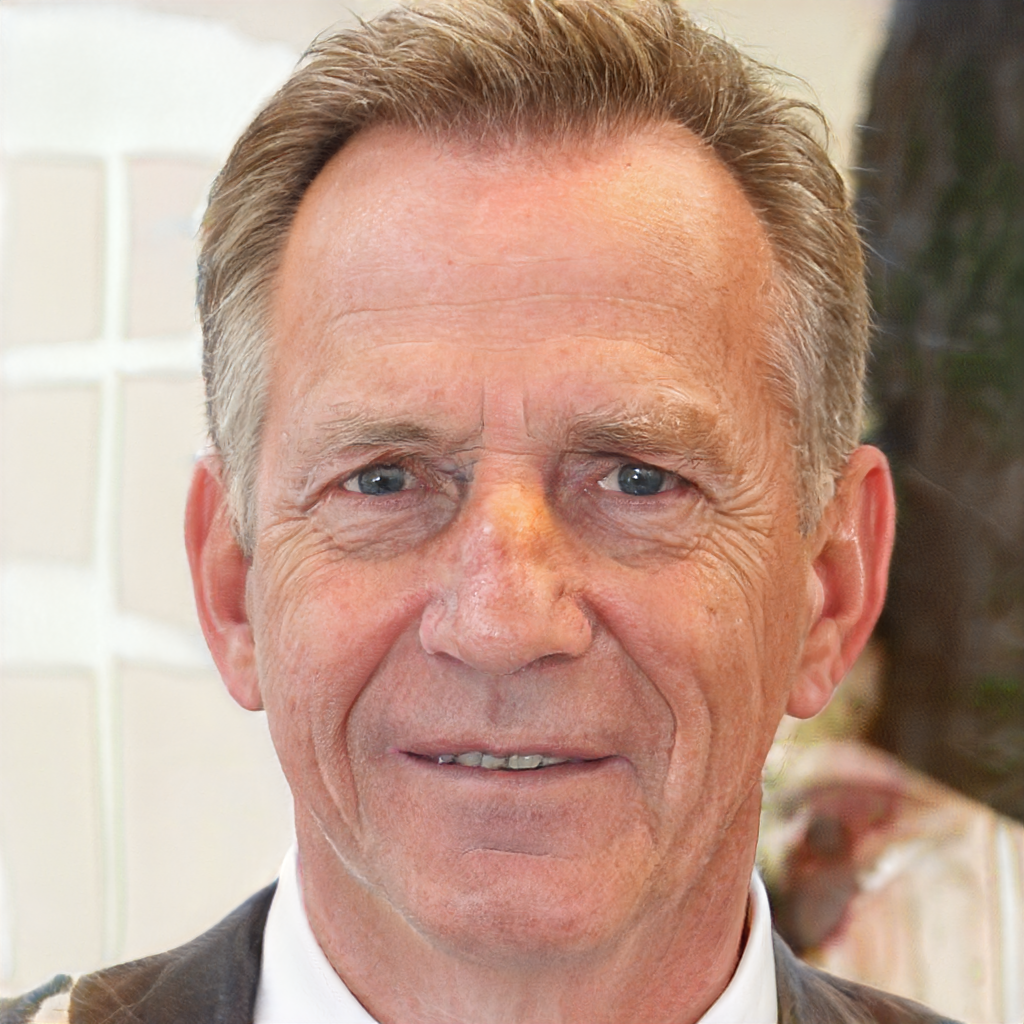}  
\end{tabular}%
}
\caption{\textbf{Attribute modification on a sample image generated from StyleGAN.} The generated images by all models exhibit successful changes in the manipulated attributes while maintaining the overall coherence of the image. Our findings indicate that the performance of \our method is comparable to state-of-the-art flow-based models in producing effective attribute manipulation.}
\label{faces}
\vspace{-2em}
\end{figure}

\paragraph{Related models:} 

The proposed simplification of the StyleGAN latent space utilizes an AutoEncoder architecture, offering advantages over flow-based models like PluGeN \cite{wolczyk2022plugen} and StyleFlow \cite{styleflow}.

Unlike PluGeN and StyleFlow, which rely on complex architectures for reversible transformations due to challenges posed by flow models, the AutoEncoder is simpler and computationally more efficient. With only two components - Encoder and Decoder, AutoEncoders facilitate easier training and demand fewer computational resources than flow-based models. Additionally, AutoEncoders autonomously learn a disentangled data representation without explicit supervision \cite{cha2022disentangling}, enhancing adaptability. In contrast, flow-based models often require intricate objective functions, posing challenges in optimization. AutoEncoders also excel in scenarios with limited labelled data, being effectively trainable with smaller datasets.

It's crucial to emphasize that StyleAE presents a unique and innovative approach, benefiting from its inherent simplicity and ease. Its architecture, based on AutoEncoders, allows for task-specific cost functions independent of distribution assumptions, contributing to a straightforward and efficient design.

\begin{wraptable}{r}{0.49\textwidth}
\vspace{-7em}
\centering
\caption{\textbf{Perceptual MSE describing the distance between embeddings of input and modified images.} We utilized an ArcFace model to extract the embedding of each image. One may observe that \our obtains significantly lower MSE than state-of-the-art models. Our qualitative experiments also demonstrate that our method preserves a substantial number of other visual facial features when manipulating just one of them.}
\bigskip
\label{L2}
\scalebox{0.85}{
\begin{tabular}{cccc}
\toprule
\textbf{Attribute}                   & \textbf{\our} & \textbf{PluGeN } & \textbf{StyleFlow} \\ \hline 
\textbf{man}     & 73.241   & \textit{\textbf{56.955}} & 53.999              \\ 
\textbf{woman}      & 36.650  & 28.300 & \textit{\textbf{23.507}}                   \\  
\textbf{no glasses} &    \textit{\textbf{20.532}}  & 49.325 & 32.092                    \\ 
\textbf{glasses}    &   \textit{\textbf{48.298}} &  52.424 & 57.537                    \\ 
\textbf{no beard}   &  \textit{\textbf{26.638}}  & 57.297 & 52.511                    \\ 
\textbf{beard}      & \textit{\textbf{4.099}} &  5.106 &  5.422                    \\ 
\textbf{no smile}   & 47.496 &    \textit{\textbf{36.251}} & 36.471                    \\ 
\textbf{smile}      & \textit{\textbf{13.697}} &   22.987 & 17.308                    \\ \bottomrule
\end{tabular}%
}
\vspace{-5em}
\end{wraptable}

In this section, we present the results for manipulating image attributes through the proposed \our{}. We use two publicly available datasets: \\Flickr Faces (FFHQ) \cite{karras2019stylebased}  and Animal Faces (AFHQv2) \cite{choi2020stargan}. Our method is compared to StyleFlow and PluGeN, which represent the current state-of-the-art. We compare our method regarding structural coherence, effectiveness in generating images with requested changes, and time efficiency. In all experiments, we combine the considered methods with the StyleGAN backbone. 

\subsection{Evaluation metrics}

The goal of attribute manipulation is to accurately modify designated image attributes while preserving other characteristics. To assess accuracy, we use classification accuracy from an independent multi-label face attribute classifier, trained on datasets not used in training the evaluated models.

To evaluate potential impacts on other image features, we employ three metrics: mean square error (MSE), peak signal-to-noise ratio (PSNR), and structural similarity index (SSIM). PSNR, a logarithmic-scale-modified MSE, and SSIM, assessing visible structures, offer insights, with higher values indicating better performance. Additionally, we calculate perceptual MSE (p-MSE) on image embeddings from a pre-trained ArcFace model \cite{deng2019arcface} sourced from the Python library \href{https://pypi.org/project/arcface/}{arcface}.

We intentionally avoided using the Frechet Inception Distance (FID) measure due to its unsuitability for attribute manipulation settings. FID primarily compares the distribution of generated images to real ones, which may not precisely reflect changes in specific attributes. As image attribute manipulation alters the distribution of generated data, FID scores can increase, even if image quality and diversity remain constant.

\subsection{Models implementation}

\our{} is a neural network comprising three fully connected layers in the encoder and decoder, each with 512 neurons and PReLU activation. Inputting a 512-dimensional style vector $w$ to the encoder yields a decoded projection $\hat{w}$ from the decoder. The omission of further latent vector compression aligns with flow-based models for a fair comparison.

\begin{wraptable}{r}{0.49\textwidth}
\vspace{-3em}
\centering
\caption{\textbf{Accuracy in modifying consecutive image attributes.} We assessed the classifier's predictive accuracy for a specific attribute, incorporating it in the final phase of vector modification for fair comparison across methods. Results indicate our plugin's efficiency, comparable to flow-based models in achieving the goal of attribute modification.}
\bigskip
\scalebox{0.8}{
\begin{tabular}{cccc}
\toprule
\textbf{Attribute}                        & \textbf{\our} & \textbf{PluGeN } & \textbf{StyleFlow} \\ \hline 
\textbf{man}      & 0.92        & 0.91            & \textbf{\textit{0.95}}    \\ 
\textbf{woman}      & \textbf{\textit{0.96}}  & 0.89      & 0.91                \\ 
\textbf{no glasses} & 0.74        & \textit{\textbf{0.78}} & 0.67                    \\ 
\textbf{glasses}    & 0.90        & \textit{\textbf{0.94}} & 0.88                    \\ 
\textbf{no beard}   & \textit{\textbf{0.96}}  & 0.95      &  \textit{\textbf{0.96}}       \\ 
\textbf{beard}      & \textbf{\textit{0.78}} & 0.55       & 0.67                    \\ 
\textbf{no smile}   & 0.99        & \textbf{\textit{1.0}} & 0.96                    \\ 
\textbf{smile}      & \textbf{\textit{1.0}} & \textbf{\textit{1.0}} & 0.99                    \\ \bottomrule
\end{tabular}
}
\label{accuracy}
\vspace{-4em}
\end{wraptable}

Training \our{} with the Adam optimizer at a learning rate of 0.0001 spans 100 epochs. To foster effective learning in both proper reconstruction and desirable attribute organization, we gradually increase the attribute loss weight factor from 0 to 0.3 over the initial 30 epochs.

Baseline comparisons include two popular attribute manipulation plugins: StyleFlow \cite{styleflow} and PluGeN \cite{wolczyk2022plugen}. Both rely on flow-based models: StyleFlow using CNF and PluGeN using NICE. We use publicly available checkpoints for evaluation, avoiding retraining PluGeN or StyleFlow ourselves.

\subsection{Manipulation of facial features}

\paragraph{Setup:} In the first experiment, we use the FFHQ dataset, which contains 70 000 high-quality images of resolution 1024 × 1024. All considered methods were trained on 10 000 images generated by StyleGAN. Eight attributes of these images were labelled using the Microsoft Face API (gender, pitch, yaw, eyeglasses, age, facial hair, expression, and baldness).

\begin{wraptable}{r}{0.49\textwidth}
\vspace{-3em}
\centering
\caption{\textbf{Average training and inference time.} Time required for training and inference on a single NVIDIA GeForce RTX 3080 GPU for each method. Results highlight the substantial speed advantage of our plugin over state-of-the-art flow-based models.}
\bigskip
\scalebox{0.85}{
\begin{tabular}{cccc}
\toprule
\textbf{Time}   & \textbf{\our} & \textbf{PluGeN } & \textbf{StyleFlow} \\ \hline 
\textbf{Training}\footnotemark[1] & $\sim$15 min & $\sim$30 min                          & $\sim$1.5 h                               \\ 
\textbf{Inference}\footnotemark[2] & $\sim$20 sec.                    & $\sim$5 min                          & $\sim$1 h                               \\ \bottomrule
\end{tabular}
}
\vspace{-2em}
\label{time}
\end{wraptable}

\footnotetext[1]{Average time of 1 training epoch.}
\footnotetext[2]{Average time taken by generation of 500 images.}

While the previous studies employing flow-based models utilized the Microsoft Face API for evaluating the accuracy of attribute manipulation, we decided to develop our own classification network due to alterations in the licensing of the Microsoft model. Our classifier is based on the ResNet18 architecture \cite{he2015deep} and consisted of 8 target classes aligned with Microsoft's classification system. 

Since every method can use different scales to represent the intensity of attributes being modelled, we employed an attribute classifier to apply a minimal modification to obtain the requested value of the attribute. In other words, we gradually modify the attribute until the classifier recognizes the attribute of the generated image with sufficient confidence. If we cannot obtain the requested modification, the classifier returns failure. All the metrics comparing original images with the modified ones, including MSE, PSNR and SSIM, are calculated on minimally modified images.

\paragraph{Results:} 
Sample results of attribute manipulation, presented in  \cref{modification} and \cref{faces}, suggest that \our{} correctly modifies the requested attributes while preserving the remaining characteristics of the image to a high extent. To support this conclusion with quantitative assessment, we analyze the classification accuracy shown in \Cref{accuracy} and the remaining metrics describing the difference between the original and modified images, see \cref{L2} and \cref{structural}. 

\begin{wraptable}{r}{0.49\textwidth}
\vspace{-3em}
\centering
\caption{\textbf{Structural reconstruction quality measures.} MSE and PSNR estimate absolute errors, while SSIM considers perceived changes in structural information. Our results indicate that \our maintains greater structural similarity between modifications and base images compared to state-of-the-art flow-based models.}
\bigskip
\scalebox{0.7}{
\begin{tabular}{ccccc}
\toprule
\textbf{Attribute}                   & \textbf{Measure} & \textbf{\our} & \textbf{PluGeN } & \textbf{StyleFlow} \\ \hline 
\multirow{3}{*}{\textbf{man}}        & \textit{PSNR} $\uparrow$     & 17.947    & \textit{\textbf{19.445}} & 19.607                    \\ 
                                     & \textit{SSIM} $\uparrow$    & 0.684    & \textit{\textbf{0.733}}  &   0.709                     \\ 
                                     & \textit{MSE}  $\downarrow$    & 0.138 & \textit{\textbf{0.130}} & 0.153                   \\ \hline
\multirow{3}{*}{\textbf{woman}}      & \textit{PSNR} $\uparrow$    & \textit{\textbf{20.485}}  & 19.026 &  19.576                    \\ 
                                     & \textit{SSIM} $\uparrow$     & 0.761 & 0.733  &  \textit{\textbf{0.822}}                    \\ 
                                     & \textit{MSE}   $\downarrow$   & 0.105   & 0.121 & \textit{\textbf{0.083}}                    \\ \hline
\multirow{3}{*}{\textbf{no glasses}} & \textit{PSNR}  $\uparrow$   &   \textit{\textbf{22.810}}  &  18.764 & 17.552                   \\ 
                                     & \textit{SSIM}  $\uparrow$   &    \textit{\textbf{0.830}}   &  0.733 & 0.800                    \\ 
                                     & \textit{MSE}  $\downarrow$    &   0.075  &  0.121   & \textit{\textbf{0.055}}                    \\ \hline
\multirow{3}{*}{\textbf{glasses}}    & \textit{PSNR}  $\uparrow$     & \textit{\textbf{18.790}}   &  18.210 & 16.840                    \\ 
                                     & \textit{SSIM}  $\uparrow$     & \textit{\textbf{0.731}} &   0.706 & 0.698                    \\  
                                     & \textit{MSE}  $\downarrow$   & \textit{\textbf{0.121}} & 0.125  &  0.156                     \\ \hline
\multirow{3}{*}{\textbf{no beard}}   & \textit{PSNR} $\uparrow$    &  \textit{\textbf{19.433}}  & 20.389 & 28.463                    \\ 
                                     & \textit{SSIM} $\uparrow$    &   \textit{\textbf{0.763}}  &  0.728 &  0.706                    \\
                                     & \textit{MSE} $\downarrow$     & 0.108  &  0.110 & \textit{\textbf{0.051}}                   \\ \hline
\multirow{3}{*}{\textbf{beard}}      & \textit{PSNR} $\uparrow$  &  \textit{\textbf{21.425}} &  20.259 &  20.769                    \\
                                     & \textit{SSIM}  $\uparrow$   &   \textit{\textbf{0.798}} &  0.769 & 0.713                    \\
                                     & \textit{MSE} $\downarrow$     & \textit{\textbf{0.088}}   & 0.109 & 0.092                    \\ \hline
\multirow{3}{*}{\textbf{no smile}}   & \textit{PSNR}  $\uparrow$   & \textit{\textbf{19.869}}  &     19.108 & 19.463                    \\ 
                                     & \textit{SSIM} $\uparrow$    & \textit{\textbf{0.750}}  &   0.739 & 0.706                    \\
                                     & \textit{MSE}  $\downarrow$    & \textit{\textbf{0.104}}   &    0.118 & 0.105                    \\ \hline
\multirow{3}{*}{\textbf{smile}}      & \textit{PSNR}  $\uparrow$   &  \textit{\textbf{22.817}} & 22.474 & 22.701                    \\ 
                                     & \textit{SSIM} $\uparrow$    & 0.831 &  0.801 & \textit{\textbf{0.833}}                   \\ 
                                     & \textit{MSE}  $\downarrow$    & \textit{\textbf{0.073}}  & 0.089 & 0.076                    \\ \hline
\end{tabular}
}
\label{structural}
\vspace{-2em}
\end{wraptable}

As can be seen from \Cref{accuracy}, in most cases, \our{} obtains comparable accuracy to PluGeN and better performance than StyleFlow. Closer inspection reveals that it is more accurate at modifying beard attributes, presents very good performance on gender and smile attributes, and slightly lower scores on the glasses feature. Taking into account image differences, reported in \cref{L2} and \cref{structural}, we can observe that \our{} better preserves most of the remaining image characteristics. While the competitive approaches excel at modifying attributes with a significant impact on the image structure (such as gender), \our showcases superior performance in manipulating more subtle attributes (such as facial expressions or an addition of~the~eyeglasses). One can see the comparison of such modifications in  \cref{modification}. In the case smile attribute the results are comparable. The outcomes demonstrate that our approach can generate images with the desired changes without considerably altering other aspects of the image, as evident from  \Cref{faces}. This highlights the ability of our method to simplify the latent space and produce more meaningful and controllable images.

Moreover, our method benefits significantly from the simplicity of the AutoEncoder approach. It requires fewer parameters and less complex architecture compared to other models, making it less time-consuming and easier to train. In fact, our model achieved comparable results with state-of-the-art flow-based models with only 100~epochs~of~training. On the other hand, these models require significantly more complicated setups and longer training times, as reported in respective papers. Furthermore, our method was more efficient in generating the same number of images compared to both models, as shown in \cref{time}.

\begin{table}[]
\vspace{-2em}
\centering
\resizebox{\columnwidth}{!}{%
\begin{tabular}{cccccccccc}
\textbf{Input} & \textbf{Cat}  & \textbf{Dog} & \textbf{Wild} &  &  & \textbf{Input} & \textbf{Cat} & \textbf{Dog} & \textbf{Wild}   \\
\includegraphics[width=0.12\linewidth]{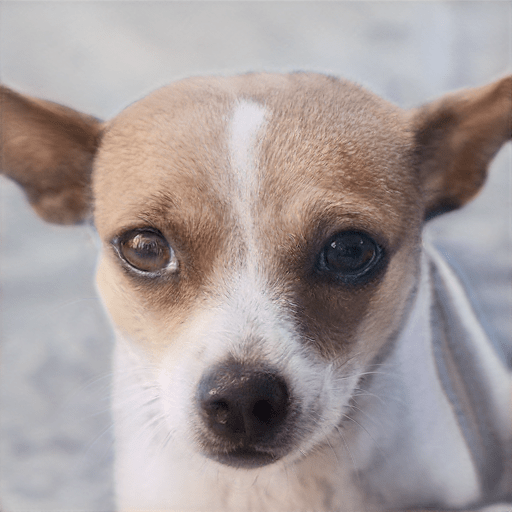} & 
\includegraphics[width=0.12\linewidth]{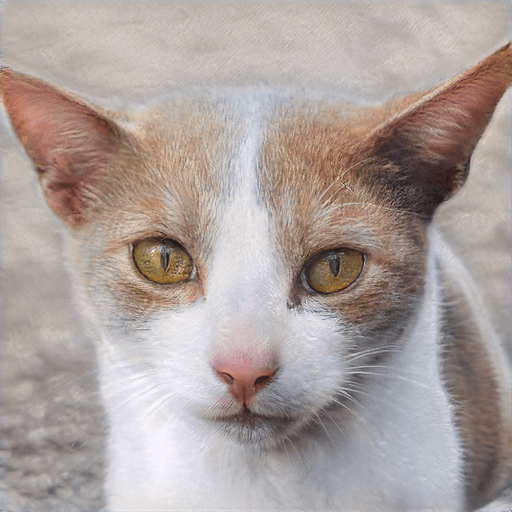} &
\includegraphics[width=0.12\linewidth]{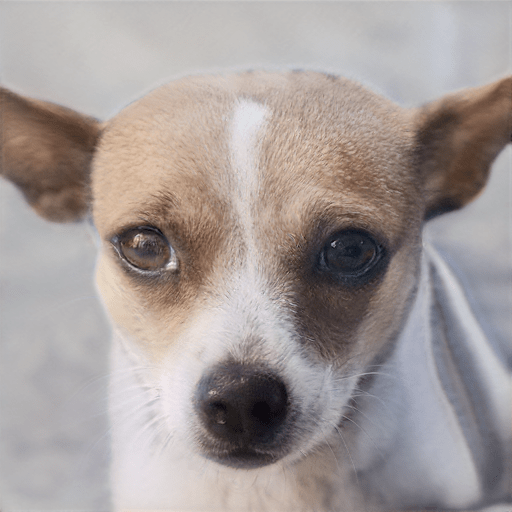} & 
\includegraphics[width=0.12\linewidth]{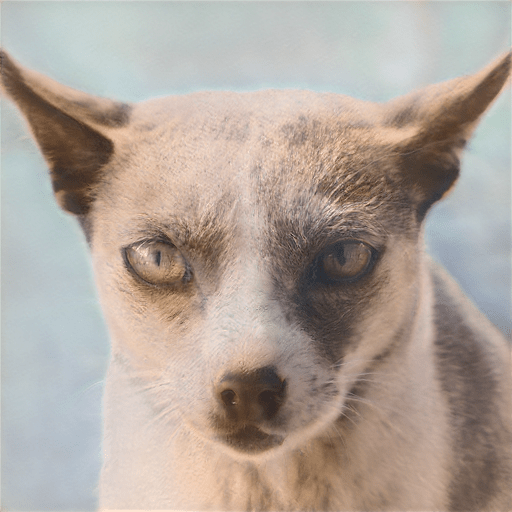} &  
 &
 &
\includegraphics[width=0.12\linewidth]{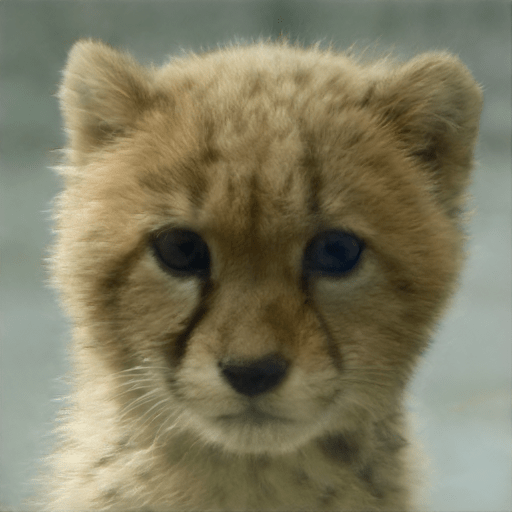} &
\includegraphics[width=0.12\linewidth]{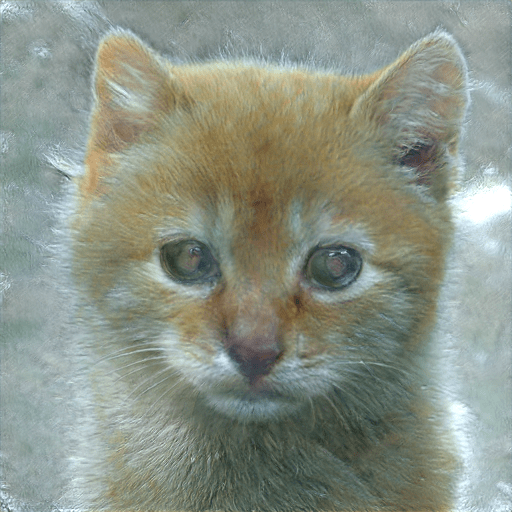} &
\includegraphics[width=0.12\linewidth]{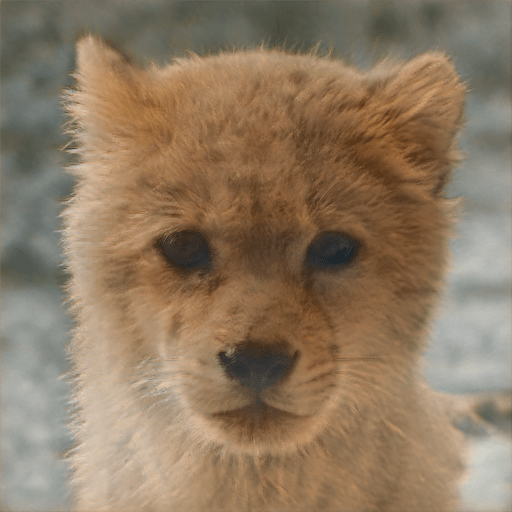} &
\includegraphics[width=0.12\linewidth]{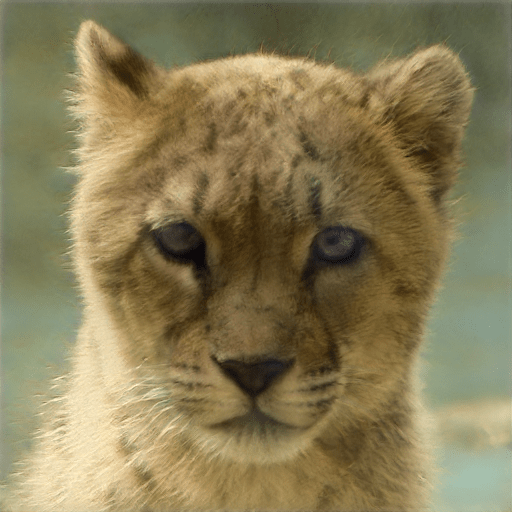} 
 \\
 \includegraphics[width=0.12\linewidth]{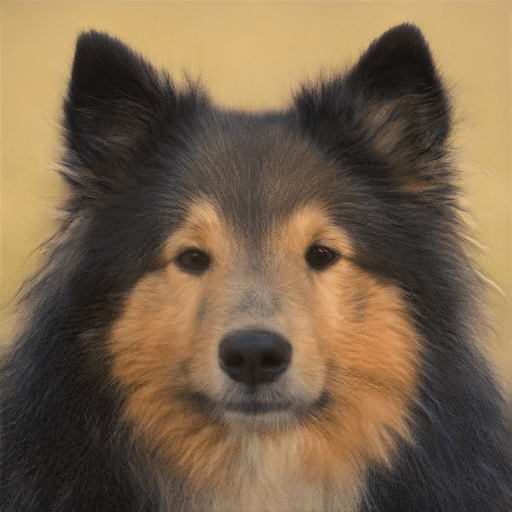} & 
 \includegraphics[width=0.12\linewidth]{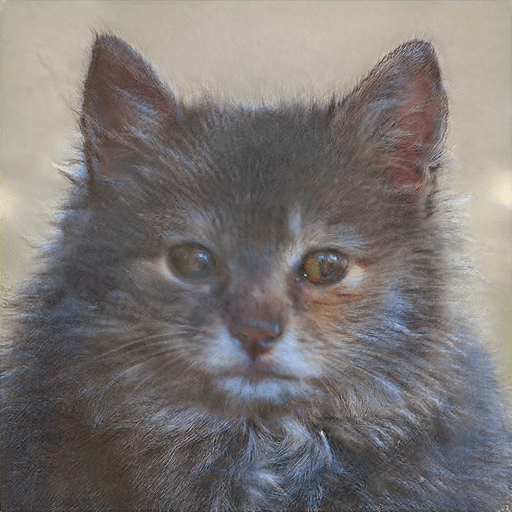} &
 \includegraphics[width=0.12\linewidth]{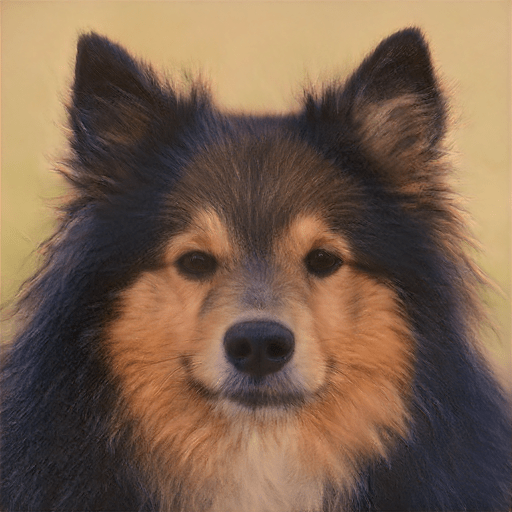} & 
 \includegraphics[width=0.12\linewidth]{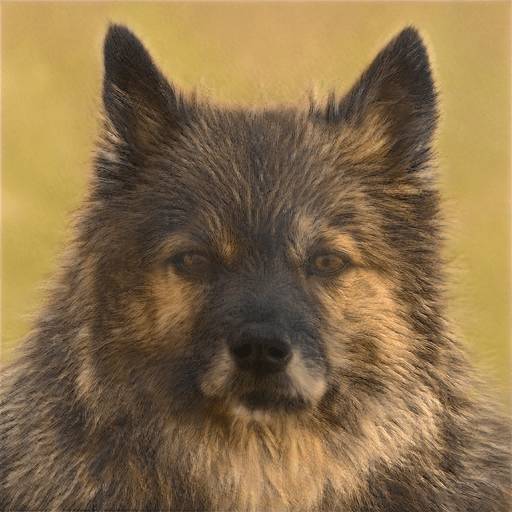} &  
 &  
 &
 \includegraphics[width=0.12\linewidth]{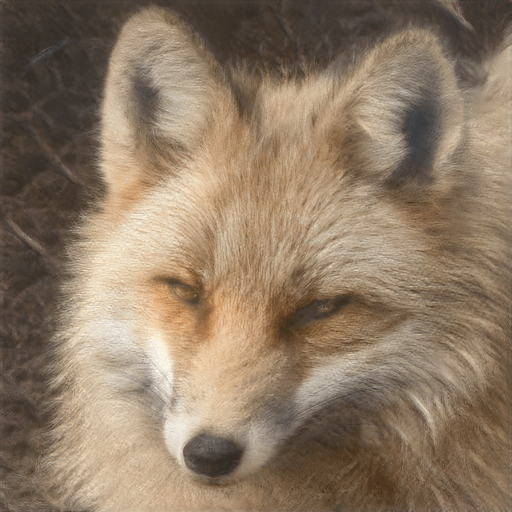} &
 \includegraphics[width=0.12\linewidth]{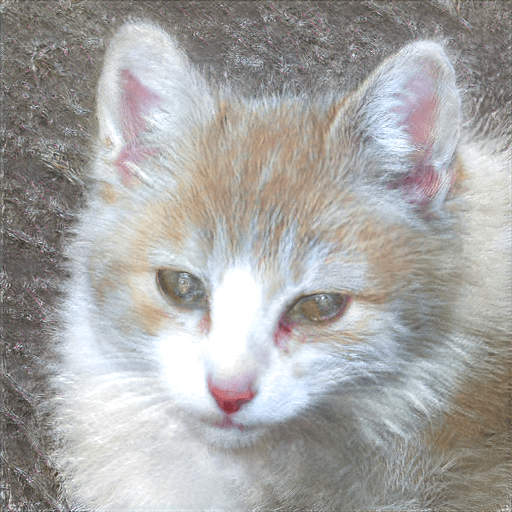} &
 \includegraphics[width=0.12\linewidth]{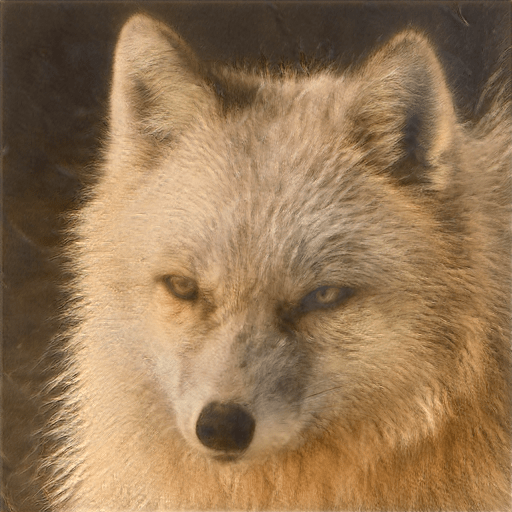} & 
 \includegraphics[width=0.12\linewidth]{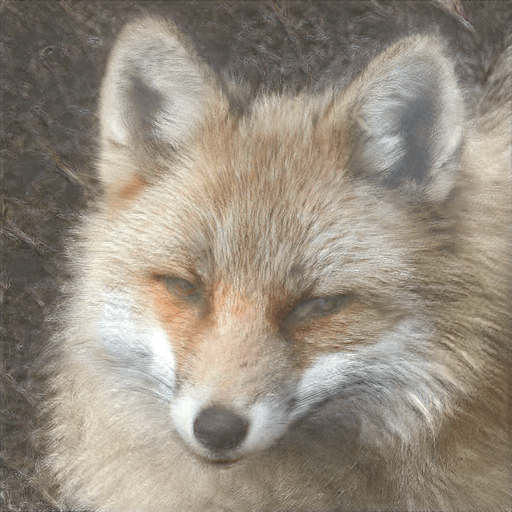}
\end{tabular}%
}
\bigskip
\caption{\textbf{Examples of attributes modification for AFHQv2 dataset.} Pose, shape and fur colour seem to be inherited from the input image. The style transfer is not ideal i.e., the quality of the input image features is not reliably copied. Finner traversing over the requested latent attribute could solve that particular issue.}
\label{animals}
\vspace{-2em}
\end{table}

\subsection{Evaluation on animal faces}

For the assessment of \our{} additional potential, a qualitative evaluation was performed utilizing the AFHQv2 dataset. This dataset comprises high-quality images featuring animal faces, categorized into specific classes, namely cats, dogs, and wild animals.

\paragraph{Setup:} 
Given the unavailability of suitable classification tools, the training of \our{} on generated images, similar to the FFHQ dataset, was unfeasible. In order to overcome this obstacle, we applied a projection technique, as presented in \cite{abdal2019image2stylegan}, to convert real images of animal faces from the AFHQv2 dataset into latent vectors of StyleGAN. These transformed vectors, marked with labels denoting the original animal class, were employed to train our model. In this specific experiment, the training of \our{} was conducted for 100 epochs, utilizing the $d_A^S(c_k, 1)$ method described in~\cref{animals-loss}.

It is essential to highlight that a direct comparison between our results and those of other methods was not feasible in this setting, because previous methods were not trained nor evaluated on the AFHQv2 dataset. Retraining the other models on the reconstructed StyleGAN latent vectors projections dataset would entail potential risks associated with the need to optimize their parameters for our specific task.

\paragraph{Results:}
In this experiment, we aimed to explore the feasibility of achieving style transfer, specifically in terms of animal type, through the modification of racial attributes.

Our empirical outcomes illustrate the effectiveness of \our plugin. This approach adeptly facilitates the transfer of specific animal classes onto generated animal faces, all while preserving integral structural characteristics like fur color and animal posture, as shown in \Cref{animals}. This outcome attests to the robustness and effectiveness of our method in producing images that conform to the desired attributes while retaining essential features.

The generated images display a high level of diversity and realism, highlighting the versatility of our approach. Notably, these results stand out considering the challenges inherent in the animal faces dataset, which encompasses a wide array of shapes and textures. 

The results of this experiment suggest that it has the potential to be applied to a variety of image-generation tasks, including those involving complex and diverse datasets.

\section{Conclusion}

This paper presents StyleAE, a novel method utilizing AutoEncoders to modify StyleGAN latent space efficiently. \our is computationally efficient and capable of generating high-quality images with controllable features across diverse datasets. 

Our experiments show that \our achieves comparable attribute modification accuracy to state-of-the-art flow-based models while being less intrusive to other image characteristics. The model's simplicity and time efficiency are key advantages.

Future research could focus on enhancing StyleAE's latent space disentanglement for more precise image control. Exploring advanced optimization methods for model fine-tuning and assessing StyleAE's efficacy in various generative model settings are promising directions.

\section{Acknowledgements}

This research has been supported by the flagship project entitled “Artificial Intelligence Computing Center Core Facility” from the Priority Research Area Digi World under the Strategic Programme Excellence Initiative at Jagiellonian University. The work of M. \'Smieja was supported by the National Science Centre (Poland), grant no. 2022/45/B/ST6/01117.

For the purpose of Open Access, the author has applied a CC-BY public copyright license to any Author Accepted Manuscript (AAM) version arising from this submission.

\bibliographystyle{splncs04}
\bibliography{ecai}

\end{document}